 \documentclass[final,1p,times]{elsarticle}

\usepackage{amssymb}
\usepackage{amsmath}
\usepackage{subcaption}
\usepackage{amsmath}
\usepackage{amssymb}  
\usepackage{makecell}
\usepackage{pifont}
\usepackage{enumitem}
\usepackage{hyperref}
\usepackage{caption}
\usepackage{booktabs}
\usepackage{multirow}
\usepackage{adjustbox}
\usepackage{float}
\usepackage{changepage}

\usepackage[table,xcdraw]{xcolor}

\newcommand{\vc}[1]{{\boldsymbol{\mathbf #1}}}
\newcommand{\lnd}{{\vc{x}}} 
\newcommand{\fix}{{F}} 
\newcommand{\mov}{{M}} 
\newcommand{\norm}[1]{\lVert#1\rVert}

\begin{document}

\begin{frontmatter}



 \cortext[cor1]{Corresponding authors}

\title{MGRegBench: A Novel Benchmark Dataset with Anatomical Landmarks for Mammography Image Registration}


\author[a]{Svetlana Krasnova\corref{cor1}} \ead{skrasnova005@gmail.com}    

\author[b]{Emiliya Starikova} 
\author[b]{Ilia Naletov} 
\author[a]{Andrey Krylov} 
\author[a]{Dmitry Sorokin\corref{cor1}} \ead{dsorokin@cs.msu.ru} 

\affiliation[a]{organization={Lomonosov Moscow State University},
            city={Moscow},
            country={Russian Federation}}
\affiliation[b]{organization={Third Opinion Platform},
            city={Moscow},
            country={Russian Federation}}    
           
\begin{abstract}

Robust mammography registration is essential for clinically relevant applications like tracking disease progression in breast tissue. However, progress has been limited by the absence of transparent public datasets and reproducible standardized benchmarks. Existing studies are often not directly comparable, as they use private data and inconsistent evaluation frameworks. To address this, we present MGRegBench, a patient-disjoint, leakage-controlled evaluation protocol for mammography registration, comprising over 5,000 image pairs, each with a breast segmentation mask, and 100 pairs with manually annotated anatomical landmarks, plus standardized train/evaluation splits and ready-to-run baselines. Using this resource, we benchmark diverse registration methods -- including classical (ANTs), learning-based (VoxelMorph, TransMorph), implicit neural representation (IDIR), a mammography-specific approach, and a recent deep learning method MammoRegNet, with implementations adapted to this modality, and validate generalization on the independent SDM-MCs dataset. Our contributions are: (1) the first public dataset of this scale with manual landmarks and masks for mammography registration; (2) a transparent, leakage-controlled benchmark enabling the first like-for-like comparison of diverse classical and machine learning-based methods; (3) external validation on SDM-MCs to test whether the main trend transfers beyond MGRegBench; and (4) an extensive analysis of deep learning-based registration. We publicly release our code and data to establish a foundational resource for fair, reproducible, and clinically relevant comparisons and catalyze future research in AI-driven medical imaging.

\end{abstract}

\end{frontmatter}




\section{Introduction}

\subsection{Clinical Motivation and Problem Statement}

Breast cancer remains the most commonly diagnosed cancer among women worldwide, accounting for 11.7\% of all cancer cases and 6.6\% of cancer deaths in 2022~\citep{international2025breast}, which tends to increase by 38\% and 68\%, respectively by 2050~\citep{kim2025global}. Early detection through mammography screening has been shown to reduce breast cancer mortality by 39\%~\citep{screeningImpact}. However, tracking disease progression and assessing longitudinal changes in breast tissue pose significant challenges due to the high variability in mammographic images acquired at different time points. Mammographic images are highly sensitive to breast compression, patient positioning, and imaging protocols. The breast is a deformable organ, and variations in compression force can lead to substantial differences in tissue appearance between examinations.

\begin{figure}[t]
  \centering
  \subfloat[2009\label{before}]{%
    \begin{minipage}{0.4\textwidth}
      \centering
        \includegraphics[width=\linewidth]{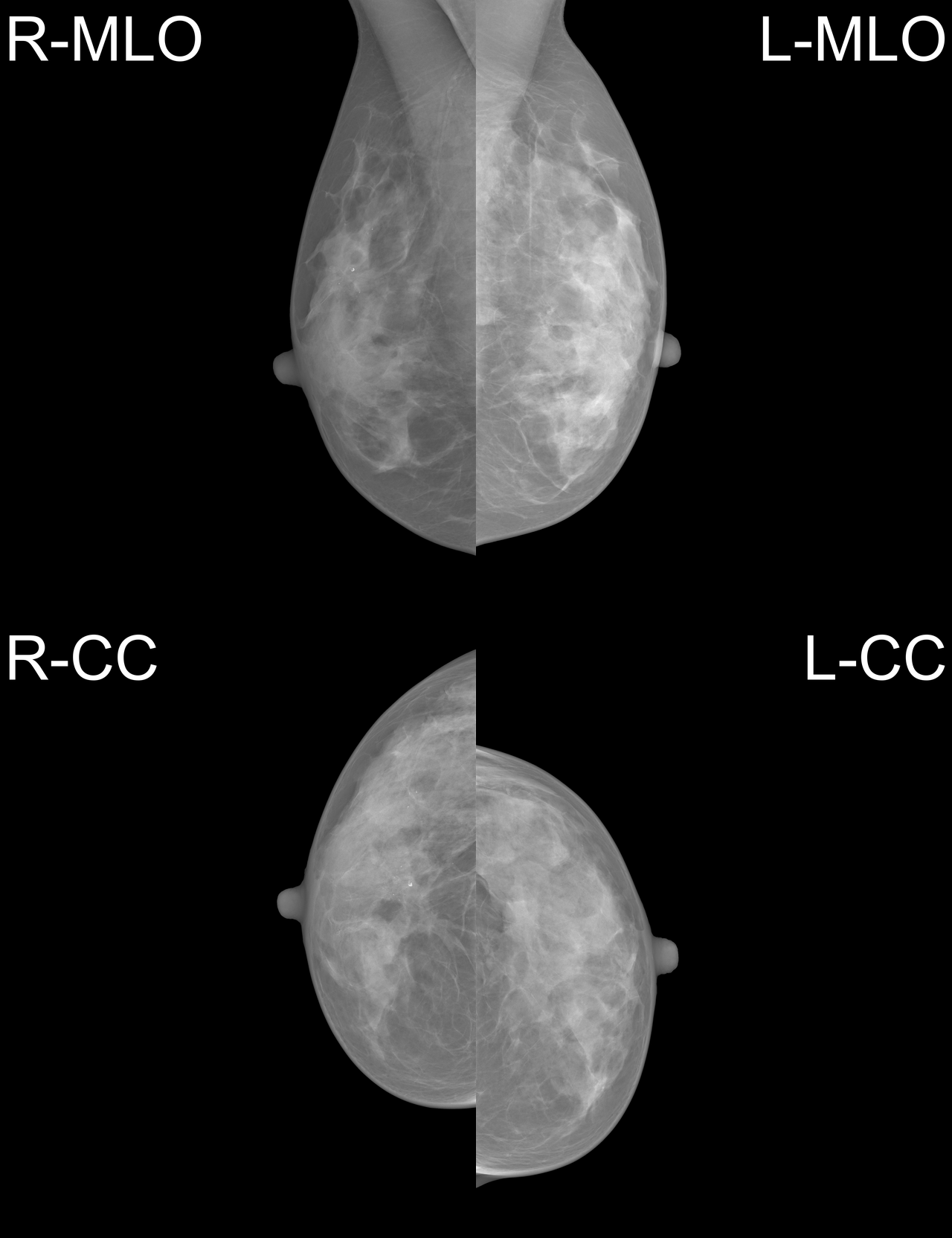}
    \end{minipage}
  }
  \subfloat[2010\label{after}]{%
    \begin{minipage}{0.4\textwidth}
      \centering
        \includegraphics[width=\linewidth]{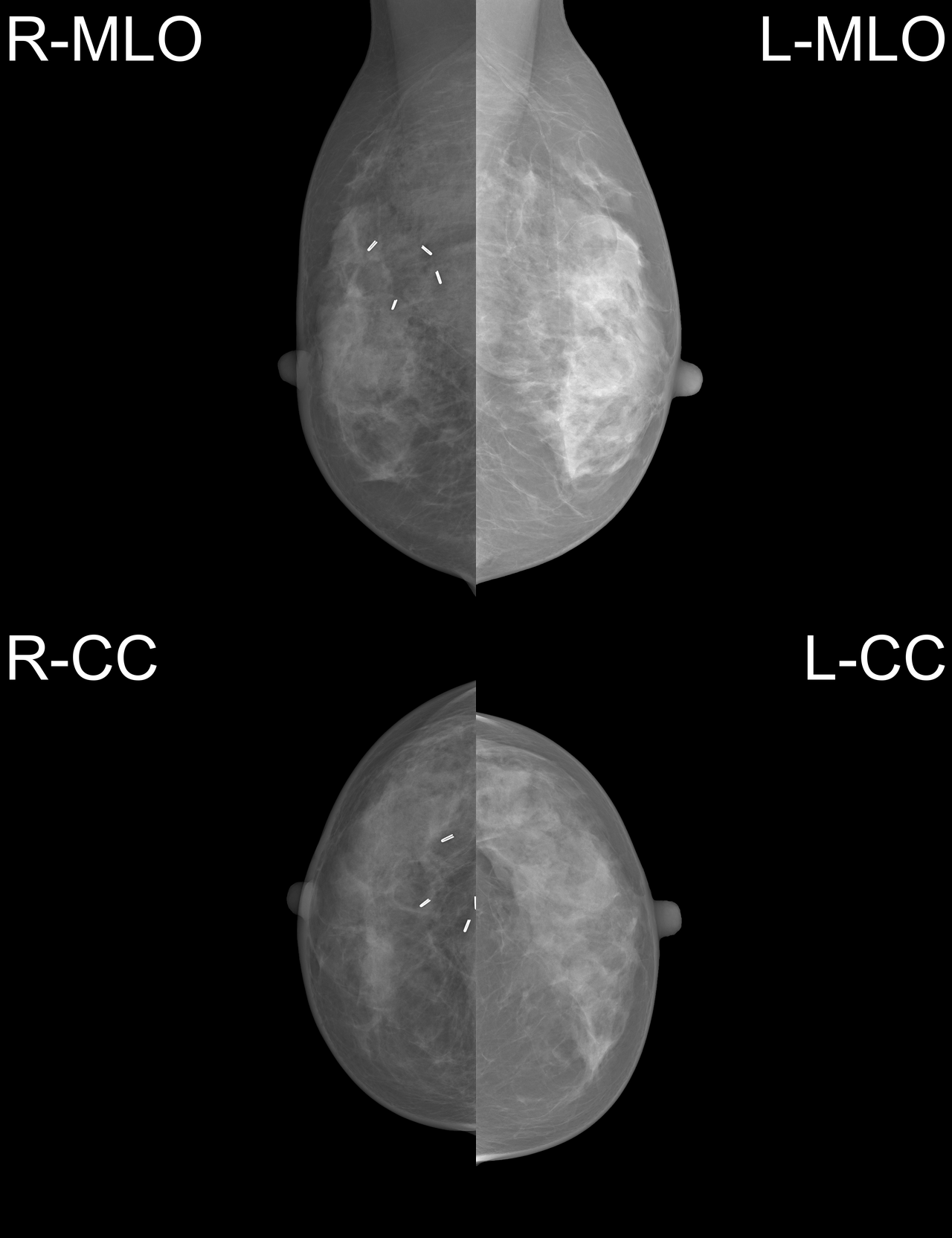}
    \end{minipage}
  }
  \caption{Example of a full-fledged paired study from INBreast dataset made in 2009 and in 2010 year correspondingly. This case is particularly illustrative due to the presence of metallic surgical clips in the right breast, indicating a prior surgical intervention.}
  \label{full-fledged-study}
\end{figure}

Recent advances in artificial intelligence have significantly enhanced the analysis of mammographic images. Deep learning methods, such as wavelet-guided CNNs, have achieved near-expert performance in breast cancer detection from mammograms~\citep{attallah2026merge}. Multi-modal approaches combining mammography with ultrasound have also demonstrated improved diagnostic accuracy by leveraging complementary information from both modalities~\citep{kvich2026multi, wang2024cnn, chen2025deep}. Furthermore, in digital breast tomosynthesis, classification methods that incorporate image quality-aware features and tumor texture descriptors have shown robustness to artifacts and noise~\citep{hassan2024classifying, lee2023transformer}. However, these works primarily focus on classification or detection tasks. The critical challenge of image registration remains underexplored, yet it is essential for tracking disease progression across time points.

To address these challenges, mammography image registration, the process of spatially aligning mammographic images, has emerged as a critical tool. Effective registration enables:
\begin{itemize} 
\item Accurate tracking of lesions (like  microcalcifications~\citep{zamir2021segmenting} or masses \citep{varlamova2024features}) over time.
\item Improved computer-aided diagnosis (CAD) by reducing false positives in longitudinal studies~\citep{shen2021artificial, mayo2019reduction, loizidou2022review}.
\end{itemize} 

The X-ray mammography registration techniques can be broadly divided into two main categories: intensity-based~\citep{IntensityBasedMethod} and feature-based~\citep{FeatureBasedMethod}. Intensity-based image registration techniques directly operate on the image pixel values. In this category, the deformation needed to register one image to the other is recovered by optimizing a measure of similarity between images. Feature-based image registration techniques require defining a set of control points that must be identified and registered in each mammography. As previously mentioned, the identification of control points is a difficult task due to the compression process that produces a nonrigid structure in which few distinct anatomical landmarks exist. Despite significant progress in deep learning-based medical image registration in recent years, very few studies have applied these techniques to mammography data~\citep{li2020mammography}. Nevertheless, a few recent works have explored deep learning approaches specifically for CC (Cranio-Caudal) - MLO (Medio-Lateral-Oblique) view registration, aiming to align complementary projections of the same breast~\citep{walton2019towards, walton2022automated}.

Despite the extensive research on mammography registration~\citep{RegMethods1, RegMethods2, thrun2025reconsidering}, a major limitation persists: the absence of publicly available, standardized datasets. Most existing studies rely on private, institution-specific data collected from individual hospitals or research centers~\citep{DatasetAbsence}. While~\citep{thrun2025reconsidering} introduced a CNN-based registration method for mammography, their evaluation on unannotated public datasets was limited. The authors omitted comparisons with existing registration techniques and assessed registration quality implicitly through the lens of a downstream clinical task and by pixel-wise image similarity metrics that do not account for anatomical accuracy.
This creates several critical challenges for the field:

\begin{itemize}
\item Lack of Reproducibility: Without access to the original datasets, independent reproduction and validation of published methods is impossible.
\item Incomparability of Methods:
Different research groups use diverse evaluation preprocessing pipelines, metrics, and datasets. Some studies report landmark registration errors~\citep{cur}, while others focus on basic metrics such as mutual information (MI) or cross-correlation (CC)~\citep{celaya2013local}.

\end{itemize}

The absence of a standardized benchmark dataset hinders progress in the field. Researchers cannot objectively determine whether new methods outperform existing ones or identify which approaches work best for specific clinical scenarios (e.g., dense breasts).

\subsection{Related Work and Existing Mammography Resources}

A number of mammography datasets have been introduced to advance research in breast cancer analysis. However, their utility for image registration task remains limited. Some datasets, such as OPTIMAM~\citep{optimam}, and LAMIS-DMDB~\citep{lamisdmdb}, are not publicly accessible without a formal data use agreement, with EMBED explicitly prohibiting the redistribution of data. Furthermore, many publicly available datasets, including DMID~\citep{dmid}, CMMD~\citep{cmmd}, VinDr-Mammo~\citep{vindrmammo}, CDD-CESM~\citep{cddcesm}, CBIS-DDSM~\citep{cbisddsm}, and MIAS~\citep{mias}, lack the paired images necessary for mammography registration task. While the large-scale Mammo-Bench~\citep{mammobench} dataset aggregates several public collections, its composite structure necessitated a detailed inspection of each constituent dataset individually to verify the availability of paired examinations from the same patient.

\begin{figure*}[!ht]
  \centering
  \subfloat[INBreast, R-MLO projection\label{short-a}]{%
    \begin{minipage}{0.3\linewidth}
      \centering
        \includegraphics[width=1\linewidth]{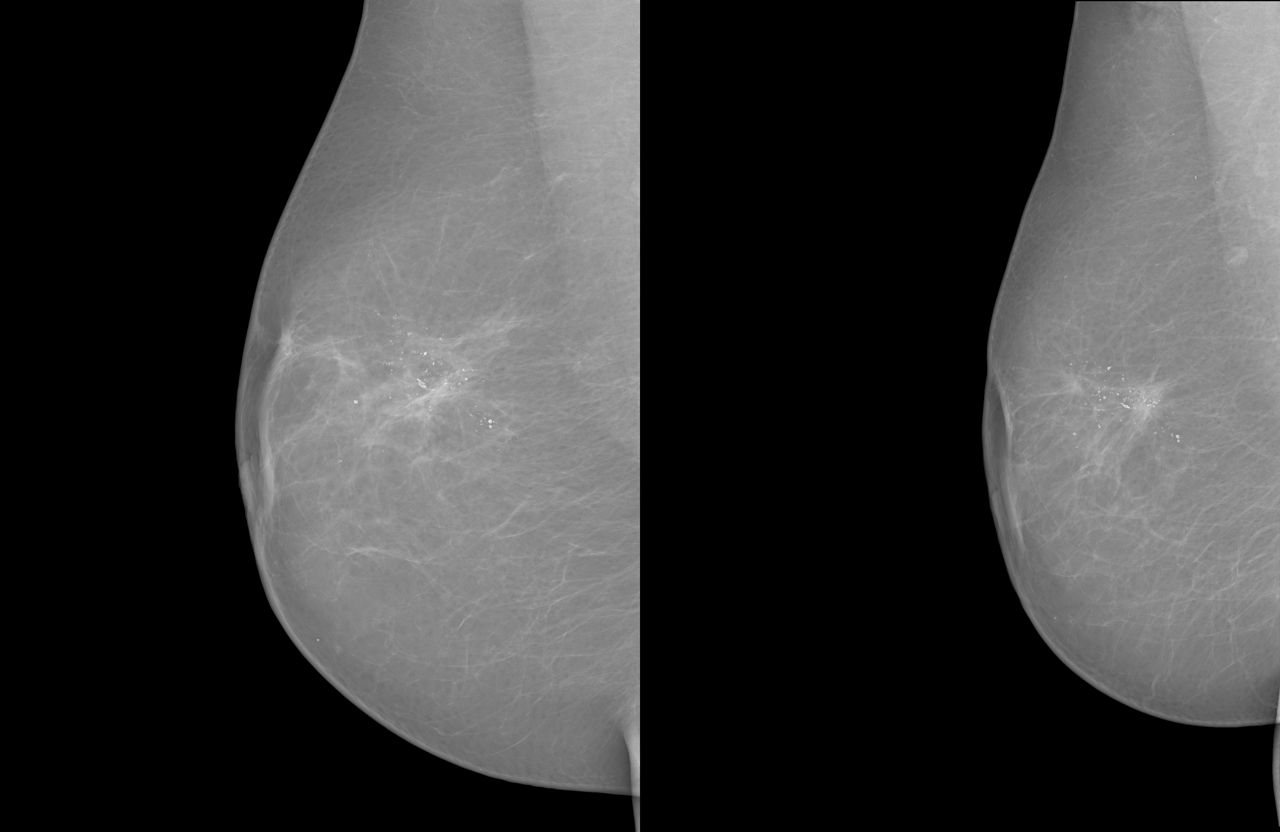}
    \end{minipage}
  }
  \hfill
  \subfloat[KAU-BCMD, R-CC projection\label{short-b}]{%
    \begin{minipage}{0.3\linewidth}
      \centering
        \includegraphics[width=1\linewidth]{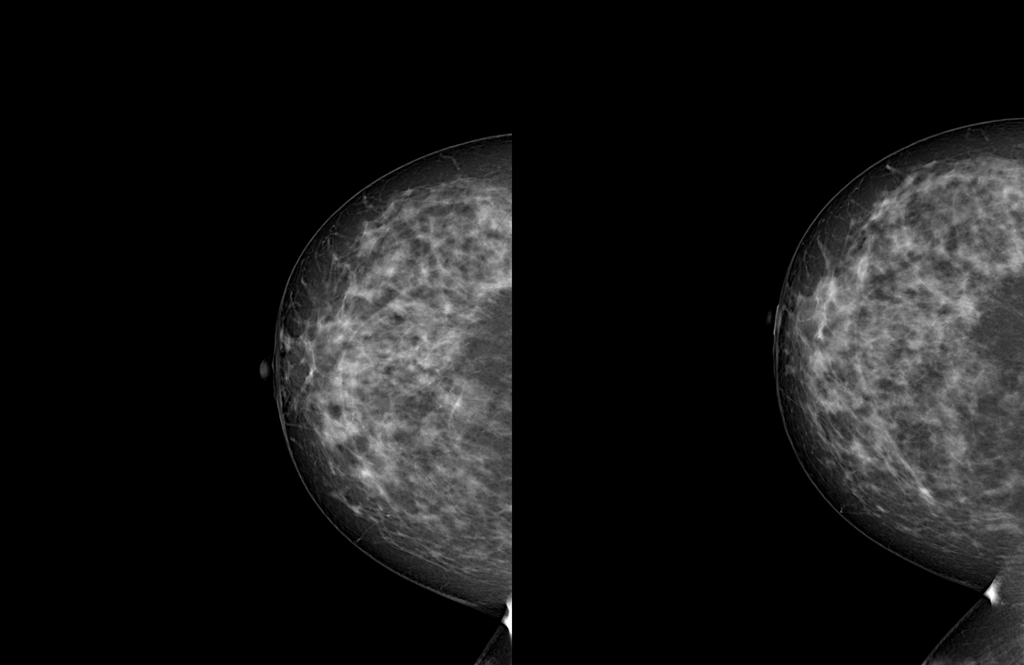}
    \end{minipage}
  }
  \hfill
  \subfloat[RSNA, L-MLO projection\label{short-c}]{%
    \begin{minipage}{0.3\linewidth}
      \centering
        \includegraphics[width=1\linewidth]{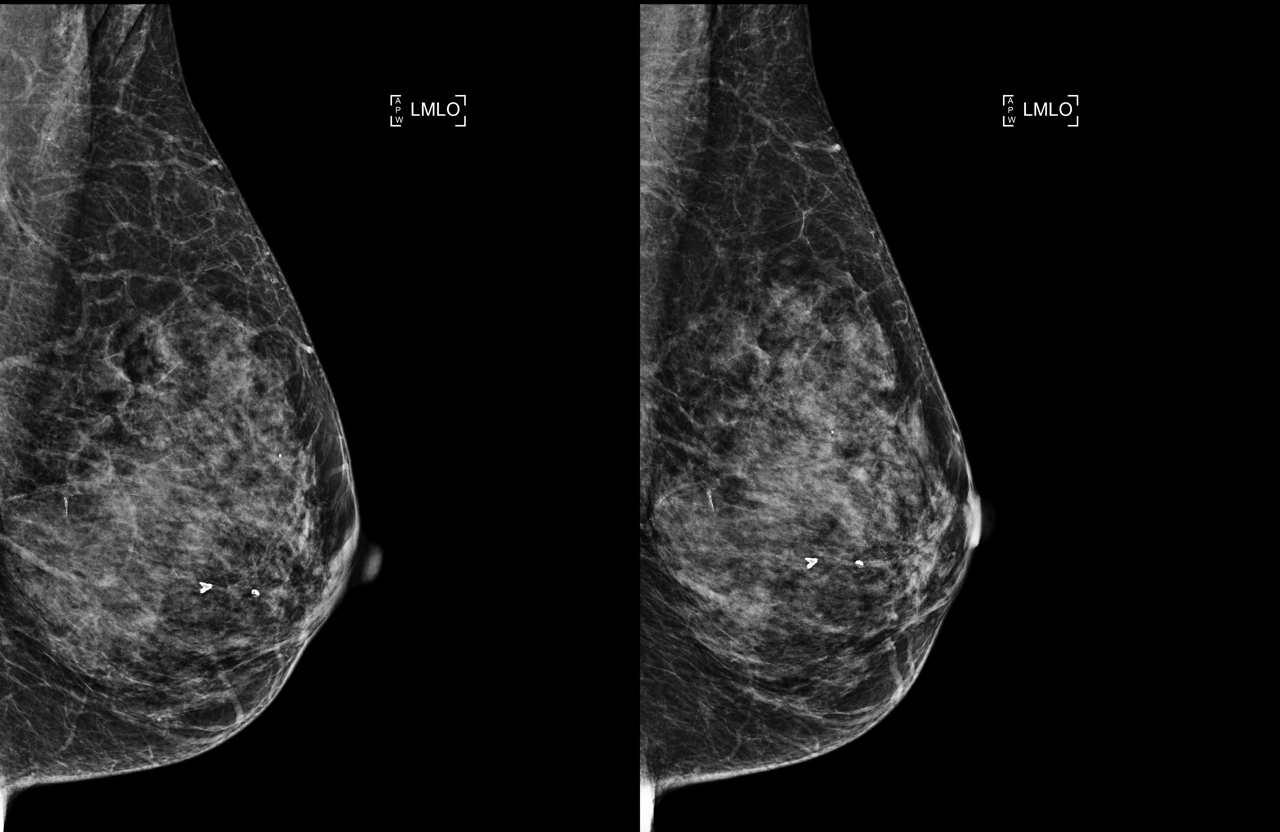}
    \end{minipage}
  }
  \caption{Example of a registration-ready image pair from each source dataset included in MGRegBench: (left) INBreast, (center) KAU-BCMD, and (right) RSNA. Each pair shows the same breast (left or right) in the same projection (CC or MLO) acquired during separate screening exams.}
  \label{dataset}
\end{figure*}

In contrast to the datasets mentioned above, 3 publicly available collections include a number of patients with multiple screening exams. From these, we extracted image pairs corresponding to \textit{the same breast} (left or right), \textit{the same projection} (CC or MLO), and \textit{the same patient}, ensuring consistent anatomical content and projection geometry suitable for pairwise registration. In many cases, these pairs correspond to different screening visits, making them valuable for longitudinal registration tasks.

\textit{INbreast}~\citep{InBreast} was introduced in 2011. It contains 115 cases, totalling 410 digital images in DICOM format. 90 cases were women with both breasts affected, each with four images, while 25 cases involved mastectomy patients; they only had two images per patient. This dataset provides detailed metadata, including patient ID, laterality, view, acquisition date, file name, breast density, BI-RADS scores, pixel-level lesion contours validated by two experts, and radiology reports.

\textit{KAU-BCMD}~\citep{KAU-BCMD}, the King Abdulaziz University Breast Cancer Mammogram Dataset consists of 1,416 cases and 2,206 images in JPG format, collected between April 2019 and March 2020. Metadata includes study date, patient ID, patient age, laterality, view, breast density, and BI-RADS scores (determined by 3 radiologists’ majority vote).

\textit{RSNA}~\citep{RSNA}, the Radiological Society of North America Screening Mammography Breast Cancer Detection dataset features a large collection of 54,705 DICOM format images from approximately 20,000 patients. Metadata includes patient ID, image ID, laterality, view, patient age, breast density, BI-RADS score, case difficulty level, the presence of breast implants, cancer (whether or not the breast was positive for malignant cancer), biopsy (whether or not a follow-up biopsy was performed on the breast), and invasive (if the breast is positive for cancer, whether or not the cancer proved to be invasive). This screening dataset does not include segmentation or ROI annotations.

Several longitudinal mammography resources have recently enabled temporal risk prediction and lesion-centered analysis, including large screening repositories such as EMBED~\citep{embed} and CSAW-CC~\citep{dembrower2020multi}, as well as sequential lesion-focused datasets such as SDM-MCs~\citep{loizidou2021breast} and SDM-Masses~\citep{loizidou2024breast}. These datasets are valuable complementary resources. However, they differ in access conditions, redistribution rights, annotation focus, and intended downstream tasks. For example, EMBED is highly valuable for large-scale mammography research but cannot be redistributed as part of an open benchmark package under its data-use restrictions. MGRegBench is therefore not intended as an exhaustive aggregation of all eligible longitudinal mammography collections, but as a controlled benchmark instance with a fixed, auditable evaluation protocol. Specifically, we release image-pair construction procedures, breast masks, expert anatomical landmark annotations, train/evaluation splits, baseline implementations, and evaluation code as a unified package. This design enables like-for-like comparison of registration methods while limiting hidden variability in preprocessing, annotation policy, and model selection. We use SDM-MCs as an independent external validation cohort rather than as a source for MGRegBench training, model selection, or benchmark construction. Other longitudinal datasets, including CSAW-CC and SDM-Masses, remain natural candidates for future MGRegBench extensions. Leakage control is addressed through explicit same-patient, same-breast, and same-projection pair construction and a frozen expert-annotated evaluation set kept separate from training and hyperparameter selection.

\subsection{Contributions}

To address these challenges, we introduce MGRegBench, the first public benchmark dataset specifically designed for mammography registration. Our main contributions are:

\begin{itemize}
\item The dataset comprises paired mammograms from multiple patients, each with an associated breast segmentation mask. A subset of 100 pairs is further annotated with expert-identified anatomical landmarks.

\item We establish a standardized evaluation framework with consistent metrics and protocols to enable fair comparison of different registration methods.

\item We provide comprehensive results by evaluating several widely used registration algorithms spanning a spectrum of algorithmic paradigms on our benchmark, creating the first standardized performance comparison in this domain.

\item We evaluate the same baselines on the independent SDM-MCs cohort to assess whether the main MGRegBench trends generalize beyond the benchmark.

\item  We provide a comprehensive benchmark of modern deep learning–based registration methods, all adapted to 2D and integrated into a single public codebase, enabling not only rigorous evaluation on mammography but also straightforward application to other 2D medical imaging modalities.

\end{itemize}

By providing this resource to the research community, we aim to accelerate progress in mammography registration, enable reproducible research, and facilitate the development of more robust clinical tools for longitudinal breast cancer analysis.

\section{Materials and Methods}

\subsection{Source Datasets and Pair Construction}

Our MGRegBench dataset is composed of mammographic images sourced from 3 publicly available datasets: INbreast~\citep{InBreast}, KAU-BCMD~\citep{KAU-BCMD}, and RSNA~\citep{RSNA} (image pairs examples are shown in Figure~\ref{dataset}).

While INbreast and KAU-BCMD include study dates, allowing us to form temporally ordered image pairs, the RSNA dataset does not provide this information. Consequently, in RSNA we can associate images only by patient ID and cannot determine the temporal order or the time interval between studies. Each MGRegBench image pair is accompanied by comprehensive metadata derived from the original datasets.

A complete clinical study for a breast is defined as a paired CC and MLO view. For the INbreast and KAU-BCMD datasets, which provide examination timestamps, we reconstructed temporal pairs by identifying a patient's complete study (L-CC, L-MLO, R-CC, R-MLO) from one time point and aligning it with a corresponding full study from a different time point. An example of a full bilateral study acquired at two different time points is shown in Figure~\ref{full-fledged-study}.

This temporal alignment was not feasible for the RSNA dataset due to the lack of timestamp information. Consequently, for RSNA, evaluation pairs were randomly sampled only within each patient (same patient ID) from cases with multiple studies available.

To facilitate rigorous benchmarking, the dataset is randomly divided into training and registration evaluation subsets. This split ensures a sufficiently large training set for deep learning methods while retaining a representative evaluation set for unbiased evaluation.

\begin{figure}[t]
  \centering
  \includegraphics[width=1\linewidth]{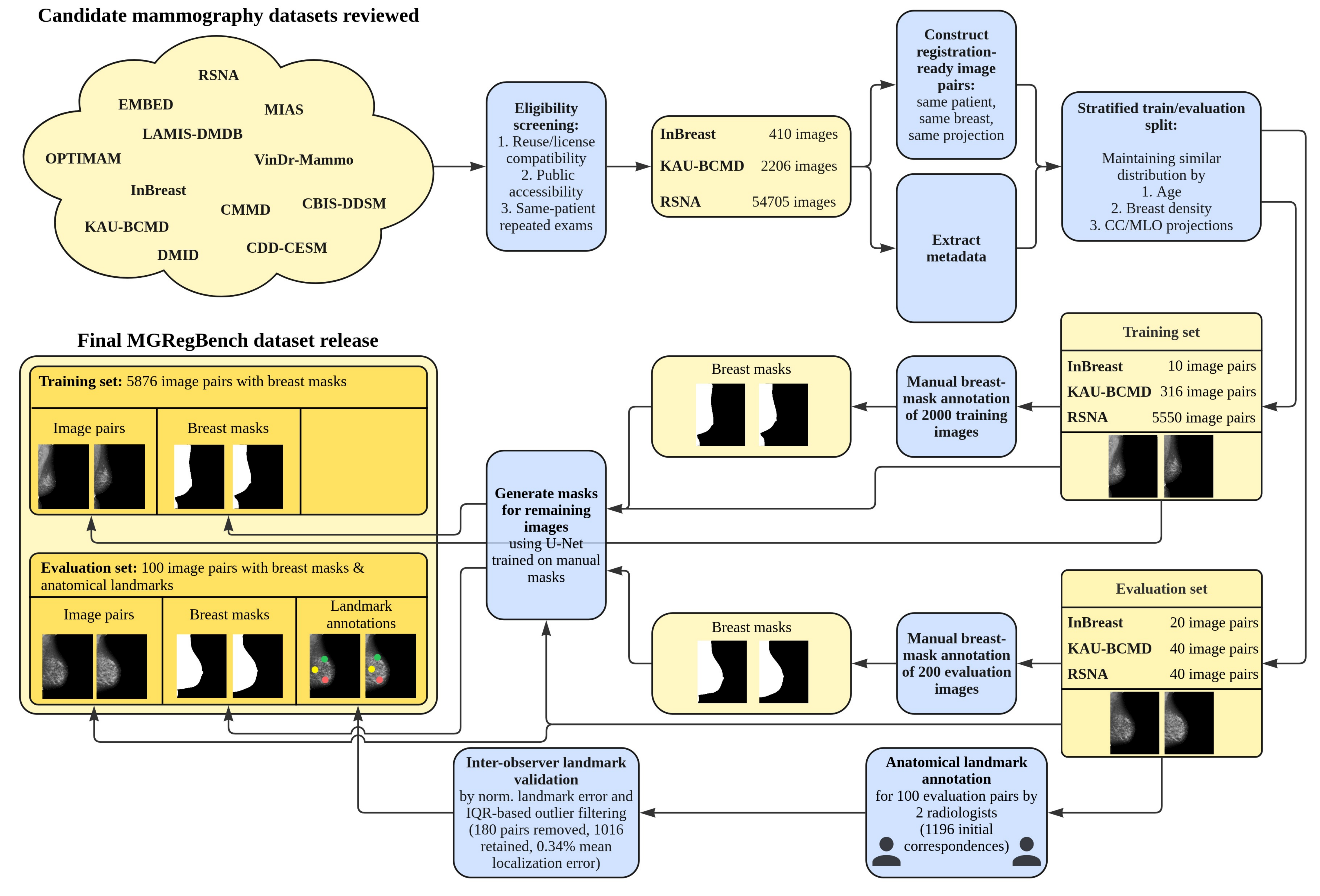}
  \caption{Data construction workflow for MGRegBench. Public mammography datasets were screened for license compatibility, accessibility, and repeated same-patient examinations. INBreast, KAU-BCMD, and RSNA were retained, harmonized, and converted into same-patient, same-breast, same-projection registration pairs. The training/evaluation split preserves similar age, density, and projection distributions. Breast masks were manually annotated and completed using U-Net predictions. Evaluation landmarks were annotated by two radiologists and filtered by inter-observer consistency, retaining 1,016 validated correspondences with a mean localization error of 0.34\%.}
  \label{fig:data_collection}
\end{figure}

The complete data construction workflow is summarized in Figure~\ref{fig:data_collection}.

\subsection{Patient-Disjoint Split and Leakage-Controlled Evaluation}

To prevent data leakage, MGRegBench uses a strictly patient-disjoint train/evaluation split. All mammograms belonging to patients selected for the evaluation subset were excluded from the training set, including all views, lateralities, and examinations. Thus, no patient-level, image-level, or landmark-level information from the evaluation set is available during training, hyperparameter selection, or model adaptation. Expert landmark annotations are used exclusively for final evaluation.

\subsection{Evaluation Set Stratification}

\begin{figure}[t]
  \centering
  \begin{minipage}{0.45\textwidth}
    \centering
      \includegraphics[width=\linewidth]{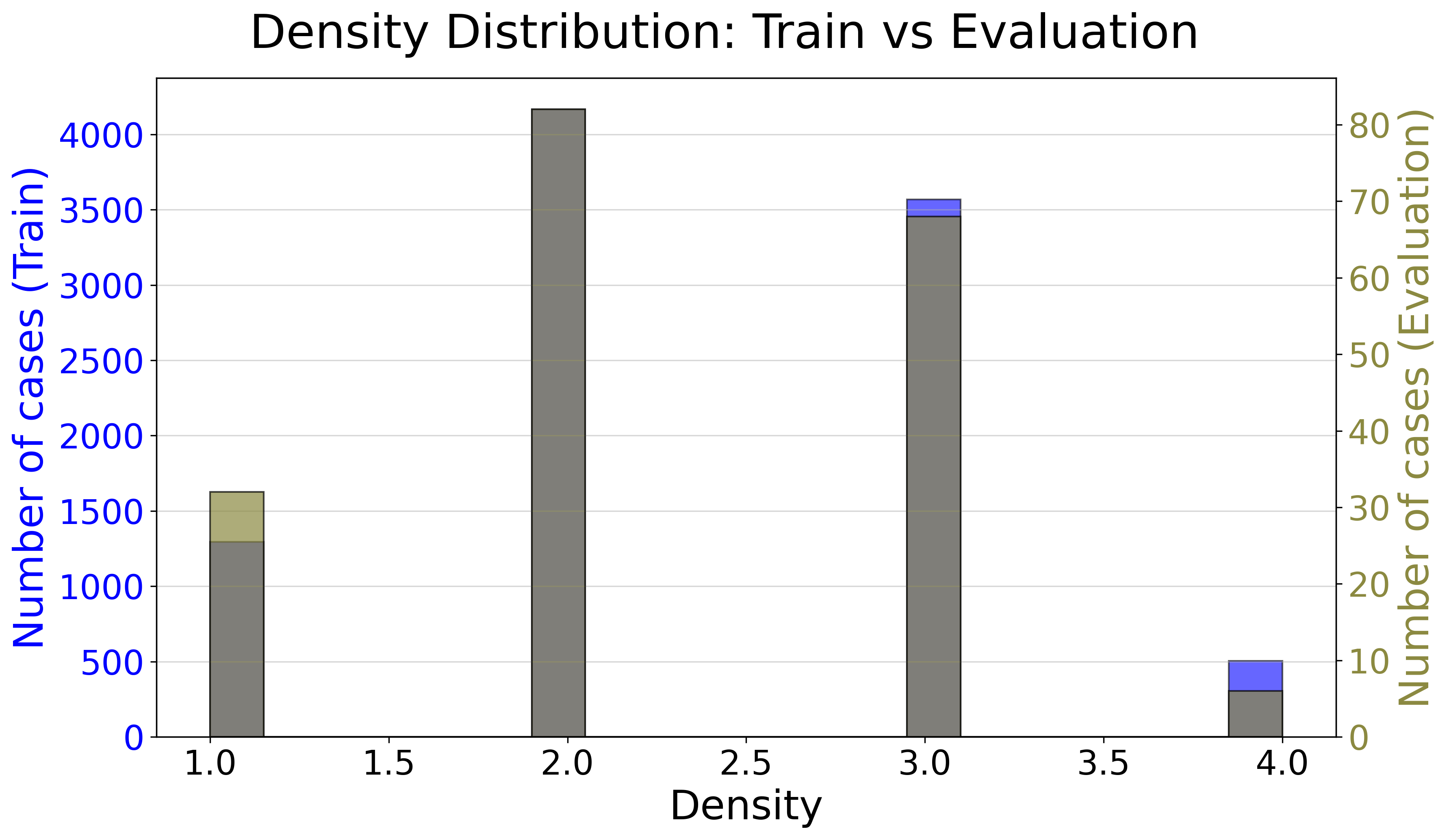}
  \end{minipage}
  \hfill
  \begin{minipage}{0.45\textwidth}
    \centering
      \includegraphics[width=\linewidth]{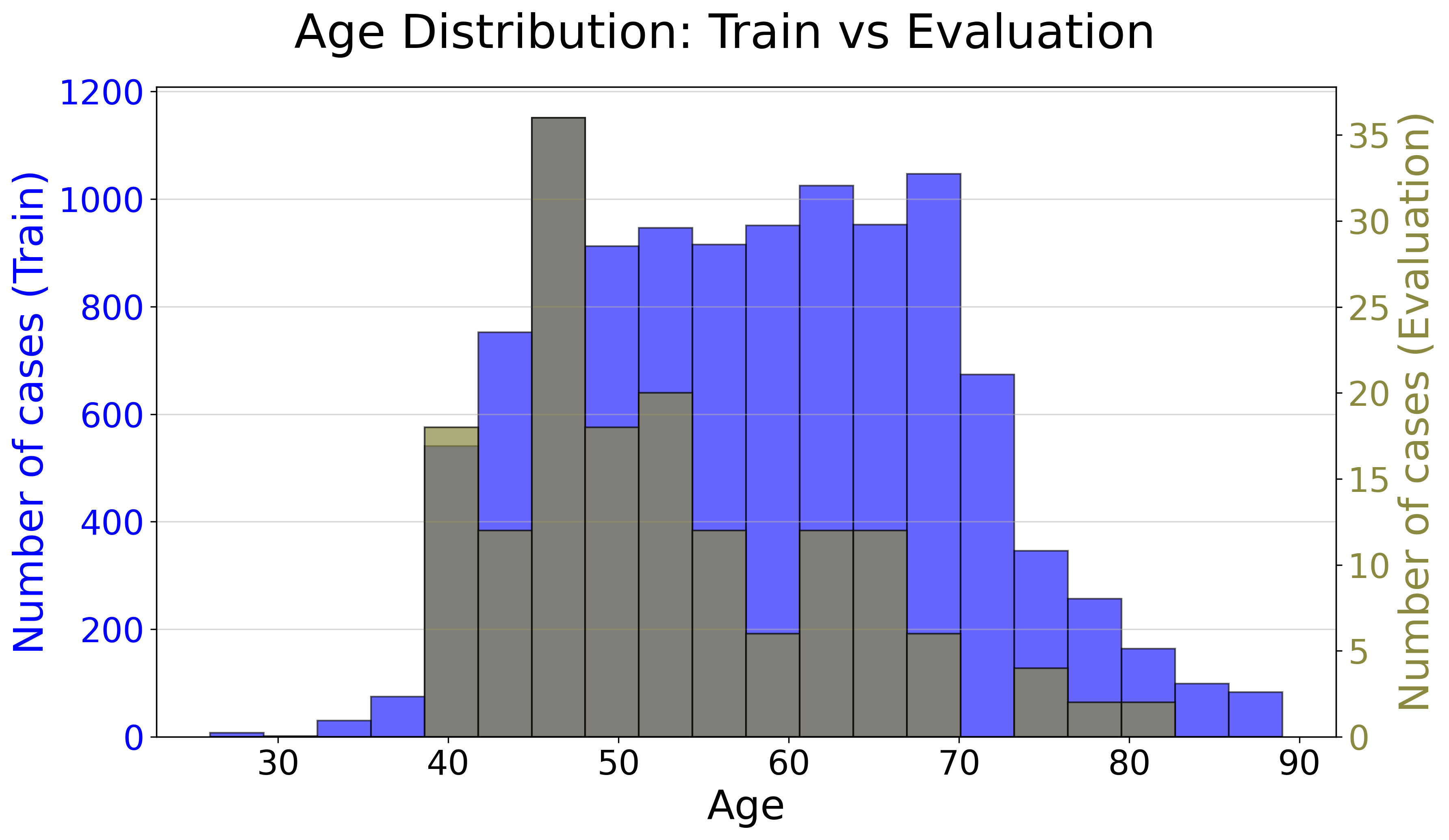}
  \end{minipage}
  \caption{Overlaid histograms of patient age and breast density distributions in the training (blue, left y-axis) and evaluation (olive, right y-axis) sets. The training histograms are plotted in the background, while the evaluation histograms are overlaid in the foreground for direct visual comparison. The x-axis represents density or age category, and the two y-axes show the respective case counts.}
  \label{metadata}
\end{figure}

The registration evaluation set was carefully selected to closely match the training set in terms of key clinical and imaging characteristics. Specifically, the distributions of breast density and patient age (Figure~\ref{metadata}) are approximately balanced between the two subsets. Breast density refers to the proportion of fibroglandular tissue relative to fatty tissue in the breast, a key radiological characteristic that influences both cancer risk and mammographic interpretability~\citep{spak2017bi}. Examples of breasts with different densities are shown in Figure~\ref{dataset}. The breast in the left image has a density of 1, while in the right it has a density of 4. The density classifications in our dataset use the original notation from the metadata (levels 1-4), which map to the A-D categories in \cite{spak2017bi}. Furthermore, the evaluation set contains an equal representation of all four standard mammographic views: R-MLO, R-CC, L-MLO, and L-CC, ensuring view-wise fairness in performance assessment.

The final evaluation subset was compiled as follows:
\begin{itemize}
    \item INbreast: 6 patients (20 image pairs). This includes 4 patients with complete bilateral studies and 2 post-mastectomy patients contributing data from a single breast across two time points.
    \item KAU-BCMD: 10 patients (40 image pairs), each providing a complete bilateral study across two time points.
    \item RSNA: 40 randomly assembled image pairs ensuring view-wise balance.
\end{itemize}

Furthermore, our evaluation set was constructed via stratified sampling to ensure proportional representation of various clinical challenges, such as varying breast density, patient age and postoperative cases. This strategy guarantees that performance metrics are not biased toward easy cases but reflect algorithm robustness across the full spectrum of real-world scenarios, making the benchmark clinically relevant.

\subsection{Anatomical Landmark Annotation and Validation}

\begin{figure}[t]
  \centering
   \includegraphics[width=0.9\linewidth]{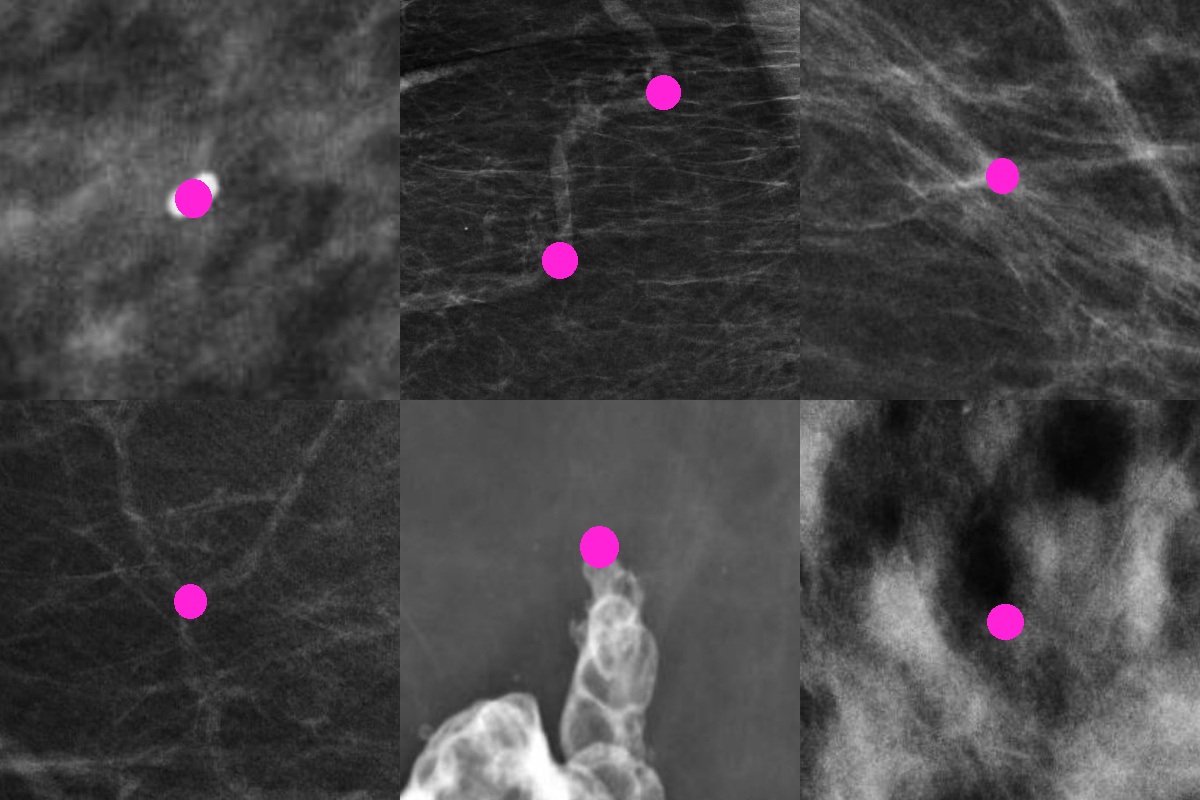}
   \caption{Examples of anatomical landmarks used for registration evaluation in MGRegBench. Each panel illustrates a distinct landmark type manually annotated by expert radiologists: microcalcifications, bends in ducts or blood vessels, vessel or duct intersections, forks, visible masses, and dark/bright blob contours.}
   \label{landmarks}
\end{figure}

Anatomical landmarks are provided exclusively for the evaluation subset of MGRegBench, enabling quantitative evaluation of registration accuracy. These landmarks include calcifications, duct or blood vessel bends, intersections, forks, visible masses and dark or bright blobs (Figure~\ref{landmarks}). 

Annotations were generated using CVAT tool~\citep{CVAT} in two phases by two professional radiologists. Annotation protocols for both phases, including an example of what constitutes corresponding landmark locations, are available \href{https://github.com/KourtKardash/MGRegBench/blob/main/Protocol.pdf}{online}. 

\begin{figure}[t]
  \centering
   \includegraphics[width=0.9\linewidth]{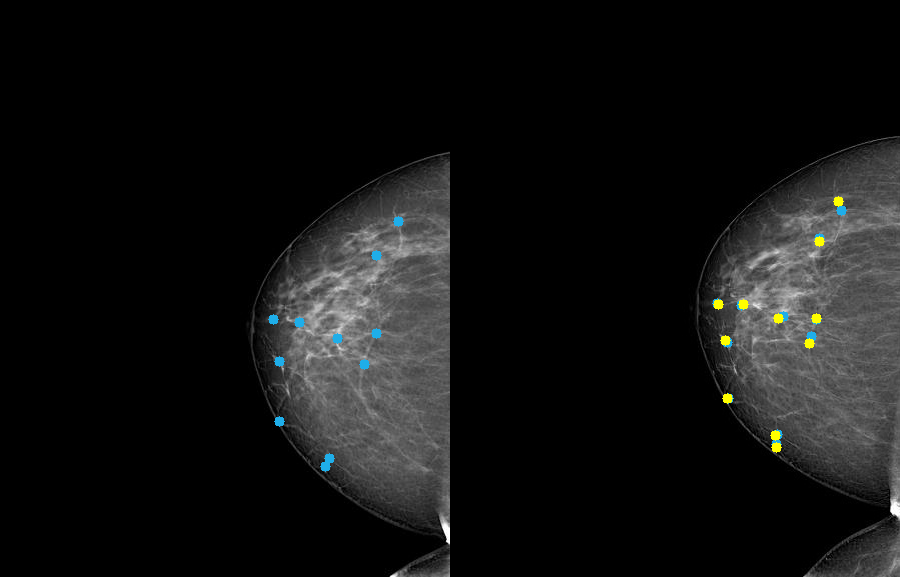}
   \caption{Example of landmark annotation performed by two radiologists. Blue markers indicate landmarks placed during the initial annotation phase; yellow markers show the refined annotations from the second (validation) phase.}
   \label{landmarks-full}
\end{figure}

During the first phase, the first annotator was asked to place 12-15 corresponding landmarks in a mammography pair that was displayed side-by-side in CVAT. This initial annotation of 100 image pairs required 31.6 hours of radiologist time. For the second phase, landmarks in the left image were fixed in place and displayed, whereas landmarks in the right one were deleted. The second annotator was then tasked with determining landmarks on the right side to the locations that they considered to match the displayed landmarks on the left side. This crucial validation phase required even more meticulous effort, taking approximately 63 hours to complete for the same set of 100 pairs. An example of landmark annotation is shown in the Figure~\ref{landmarks-full}.

We computed the localization error as the Euclidean distance between corresponding landmark locations annotated independently by two expert radiologists. To account for varying image resolutions, the error was normalized by corresponding image diagonal. Outliers were identified using the interquartile range ($IQR$) method: a normalized error was considered an outlier if it exceeded the threshold $T$ defined as
\begin{equation}
    T =Q_3 + 1.5IQR,
\label{iqr}
\end{equation}
where $Q_1$ and $Q_3$ denote the 25th and 75th percentiles, respectively, of the normalized localization error distribution computed over all annotated landmark pairs, and $IQR=Q_3-Q_1$.

\subsection{Breast Mask Annotation and Segmentation}

We manually annotated binary breast masks for the entire evaluation subset and for 2,000 images from the training set. Using these high-quality annotations, we trained a U-Net segmentation model \citep{ronneberger2015u} with a composite loss function that combines binary cross-entropy (BCE) and Dice loss:

\begin{equation}
    L = 0.5\cdot BCE + 0.5\cdot(1 - Dice),
\label{unet}
\end{equation}
optimized with a learning rate of $10^{-4}$ and batch size of 4 for 50 epochs.

As a result, every image in MGRegBench is provided with a corresponding binary mask. To support reproducibility and future work, we publicly release the trained segmentation model weights.

\subsection{Dataset Structure, Access, and Licensing}

The MGRegBench dataset is organized into \textit{train} and \textit{evaluation} directories for method development and performance assessment, respectively.

Data in both directories are grouped by source datasets (\textit{INBreast}, \textit{KAU-BCMD}) within patient-specific folders named with anonymized IDs. Each patient folder contains 2 (or more in some cases) images: PNG (and original DICOM) for INBreast and JPG for KAU-BCMD.

Due to licensing, the RSNA dataset is not redistributed. Instead, we provide a preprocessing script that converts a local copy of the official RSNA data into the MGRegBench format (PNG images with preserved DICOM metadata), ensuring full reproducibility.

Ground-truth breast segmentation masks are in the \textit{train-masks} and \textit{evaluation-masks} directories, mirroring the image structure. Masks for all 3 source datasets are stored in PNG format.

Expert anatomical landmark annotations for the evaluation set are in 3 separate XML files:\
(1) landmarks by the first radiologist on the first image of each pair,\
(2) corresponding landmarks by the same radiologist on the second image, and\
(3) independent annotations by a second radiologist on the second image for inter-observer validation.

MGRegBench is publicly available\footnote{\url{https://github.com/KourtKardash/MGRegBench}}.
The repository includes the benchmark project structure, representative example data, dataset preparation utilities, complete evaluation code that reproduces all metrics reported in this paper, and our mammography-specific adaptations of representative classical and learning-based registration baselines.
Dataset documentation and instructions for running the registration methods and reproducing the benchmark evaluation are provided in the repository README\footnote{\url{https://github.com/KourtKardash/MGRegBench/blob/main/README.md}}.

Code is released under the MIT and Apache 2.0 licenses.
MGRegBench redistributes data under the original licenses of the underlying sources (CC0 and CC-BY-NC-4.0).
For RSNA, we do not redistribute images; instead, we provide a script that builds the RSNA portion of MGRegBench from a locally available copy of the dataset in accordance with the RSNA data access terms.

\subsection{Registration Baselines and Selection Rationale}

For mammography image registration, we employed 6 representative methods spanning a spectrum of algorithmic paradigms:

\noindent\textbf{ANTs (SyN)~\citep{avants2009advanced, syn}} as a widely adopted classical method being standard in medical image registration;

\noindent\textbf{A curvilinear coordinate-based method~\citep{cur}} explicitly designed for breast anatomy by aligning images along tissue-structure–aware coordinate systems;

\noindent\textbf{IDIR~\citep{inr}}, a coordinate-based method using implicit neural representations (INRs), which models deformation as a continuous function for inherent smoothness and generalization.

\noindent\textbf{VoxelMorph~\citep{vxm} and TransMorph~\citep{tm}} as most common well-founded deep learning–based approaches that learn deformable registration from data, with TransMorph leveraging transformer architectures to capture long-range spatial dependencies;

\noindent\textbf{MammoRegNet~\citep{thrun2025reconsidering}} as a recent deep learning method designed for 2D mammography data. 

This selection spans the primary methodological families in registration (classical, anatomy-aware, implicit, and deep learning-based) while ensuring compatibility with our landmark-based, whole-image evaluation protocol.

\subsection{Evaluation Metrics}
To provide a comprehensive and clinically meaningful evaluation of registration performance, we employ a diverse set of metrics spanning anatomical accuracy, intensity-based similarity, segmentation-based overlap, deformation regularity, and computational efficiency.

\textbf{Anatomical accuracy}.
Anatomical accuracy is assessed using the relative Target Registration Error (rTRE), computed on expert-annotated corresponding landmarks.
The rTRE is defined as the Euclidean distance between corresponding landmarks in the transformed moving image ($I_\mov$) and the fixed image ($I_\fix$), normalized by the image diagonal length and expressed as a percentage:
\begin{equation}
    \operatorname{rTRE}_l = \frac{\norm{\lnd^{\mov}_l-\lnd^{\fix}_l}_2}{d} \cdot 100,
\label{eq:rtre}
\end{equation}
where $\lnd^{\mov}_l$ and $\lnd^{\fix}_l$ are the landmarks, $d$ is the image diagonal length.

To obtain a robust estimate of registration error with both intra- and inter-observer uncertainty in landmark placement, we employed the following evaluation protocol:\\
(1) compute the mean Euclidean distance between the first annotator’s landmarks on the moving image (after warping) and their corresponding landmarks on the fixed image (also by the first annotator);
(2) compute the mean distance between the same warped moving landmarks and the second annotator’s landmarks on the fixed image;
(3) average these two distances per image pair;
(4) finally, average the result across all pairs in the evaluation set.

\textbf{Intensity-based similarity metrics}.
Intensity-based similarity is evaluated using 4 complementary metrics: 
\begin{itemize}
    \item Mean Squared Error (MSE) measures pixel-wise intensity differences;
    \item Structural Similarity Index (SSIM) quantifies perceptual similarity;
    \item Mutual Information (MI) evaluates the statistical dependence between image intensities;
    \item Cross-Correlation (CC) assesses the linear correlation of intensity values;
\end{itemize}

\textbf{Deformation regularity metric}.
Deformation regularity is critical for clinical plausibility. The foldings of the deformation field were measured using the percentage of negative Jacobian determinants (NJD).

\textbf{Segmentation-based metric}.
Binary breast tissue masks are available for every image in the MGRegBench dataset. A moving mask is warped by the deformation field, enabling per-pair computation of segmentation-based metric after registration. Specifically, we compute the Dice Similarity Coefficient (DSC) to quantify overlap.

\textbf{Computational efficiency measurements}.
To assess practical deployment potential, we measure computational efficiency:
\begin{itemize}
    \item GPU memory (VRAM) usage during inference and training (for trainable methods).;
    \item Runtime per image pair for inference and total training time (for trainable methods).
\end{itemize}

Experiments were run on an Intel Xeon Gold 6226R CPU / NVIDIA A6000 GPU server.

Together, these metrics provide a holistic view of registration quality balancing anatomical fidelity, image similarity, and real-world usability.

\subsection{Experimental Setup and Computational Measurements}
ANTs (SyN) served as a classical, intensity-based baseline. We used the default parameters of the Symmetric Normalization algorithm without any modality-specific tuning.

The curvilinear coordinate-based method incorporates breast-specific anatomical priors to constrain the deformation field, ensuring physically plausible alignments. It is one of the prominent non-learning methods explicitly designed for mammographic registration. As the original work does not provide an official implementation, we release our re-implementation to ensure reproducibility and facilitate future research.

All deep learning methods were preceded by initial affine alignment perfomed with ANTs~\citep{avants2009advanced} to ensure robust convergence. MammoRegNet incorporates an affine alignment stage within its architecture. To assess the impact of external initialization, we report two variants: (1) MammoRegNet applied without external affine pre-alignment and (2) MammoRegNet preceded by an affine transform computed with ANTs~\citep{avants2009advanced}. We report the performance of this affine baseline alongside the full results to explicitly demonstrate the incremental improvement achieved by the more complex, non-rigid registration algorithms.

IDIR was adapted to 2D input data. For each image pair, the model was trained for 2500 epochs using a batch size of 2500 coordinate points,  with a \(10^{-4}\) learning rate, normalized cross-correlation (NCC) loss, ReLU activation, and a Jacobian deformation regularization weight of 0.01.

VoxelMorph and TransMorph were adapted to 2D and trained on the MGRegBench train set for 500 epochs (100 iterations/epoch) with a batch size of 4. Using an MSE loss, all hyperparameters were adopted from the original implementations except the deformation regularization weight, which was set to 0.5 for both methods and selected based on preliminary experiments on the training data only and fixed prior to evaluation to ensure smooth, fold-minimized deformations.

Finally, MammoRegNet was employed without any modifications and trained for 500 epochs (100 iterations/epoch) with a batch size of 4, using the hyperparameters specified by the authors.

For external validation, the same method configurations and evaluation code were applied to the SDM-MCs dataset~\citep{loizidou2021breast}, which originally contains 50 normal cases and 50 suspicious cases with BI-RADS 4-5 assessments. Each case consists of two projections (CC and MLO) acquired at two time points.

For this experiment, we selected 16 longitudinal image pairs (8 MLO and 8 CC) in which clear corresponding microcalcification locations could be established across time points. For these pairs, 48 point correspondences were annotated to evaluate registration accuracy. These curated correspondences were used only for evaluation. The trainable deep learning baselines (VoxelMorph, TransMorph, MammoRegNet, and Affine + MRN) were not retrained or fine-tuned on SDM-MCs. We used the weights learned on MGRegBench and kept the remaining baseline settings fixed.

\section{Results}

\subsection{Dataset Statistics and Annotation Quality}

The training set comprises 10,913 mammographic images, which form 5,876 image pairs for registration. This set is designed to facilitate the development and evaluation of both classical and learning-based image registration algorithms, including data-intensive deep learning models such as VoxelMorph and TransMorph. The evaluation set contains 200 mammographic images, forming 100 pairs for registration. For each of the 11,113 images in the MGRegBench dataset, we provide a corresponding binary breast segmentation mask.

The annotation of anatomical landmarks in mammography is inherently challenging due to the overlapping nature of breast tissue, which obscures clear and consistent anatomical landmarks — a difficulty that is also recognized in other areas of deformable 2D medical imaging domains~\cite{xia2018thorax}. Despite this, our dataset includes 100 expert-annotated image pairs, each labeled with approximately 10 clinically meaningful landmarks. This places MGRegBench among the most substantial publicly available 2D registration benchmarks with manual landmark annotations. For context, the FIRE fundus registration dataset includes 134 image pairs with 10 landmarks each~\cite{hernandez2017fire}; the fetal brain ultrasound benchmark provides 25 landmarks across all 104 images~\cite{cabezas2024benchmark}; and the thorax X-ray/CT registration dataset uses only 8 landmarks per image due to anatomical ambiguity~\cite{xia2018thorax}.

After applying the outlier filtering criterion, 180 out of the initial 1,196 annotated landmark pairs were excluded, corresponding to approximately 15\% of the total. The average localization error between corresponding landmark locations after filtering is 0.34\%.

The trained model was then applied to the remaining training images to generate segmentation masks automatically. On the evaluation subset, it achieves an average Dice score of 0.9959, confirming its high accuracy.

\subsection{Main Benchmark Results}

\begin{figure}[H]
  \centering
  \subfloat[L-CC projection\label{CC}]{%
    \begin{minipage}{0.8\textwidth}
      \centering
        \includegraphics[width=\linewidth]{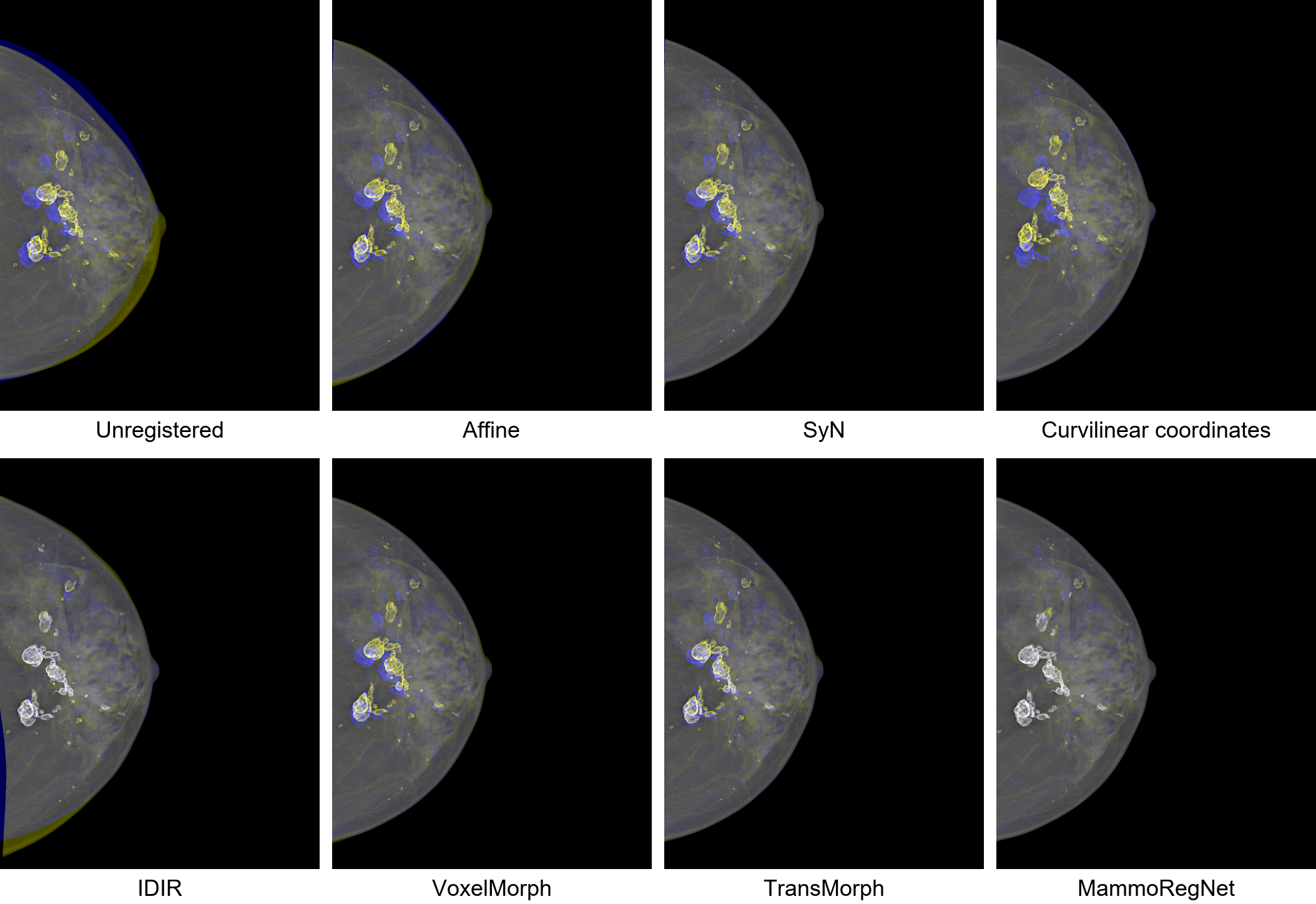}
    \end{minipage}
  }\\
  \subfloat[L-MLO projection\label{MLO}]{%
    \begin{minipage}{0.8\textwidth}
      \centering
        \includegraphics[width=\linewidth]{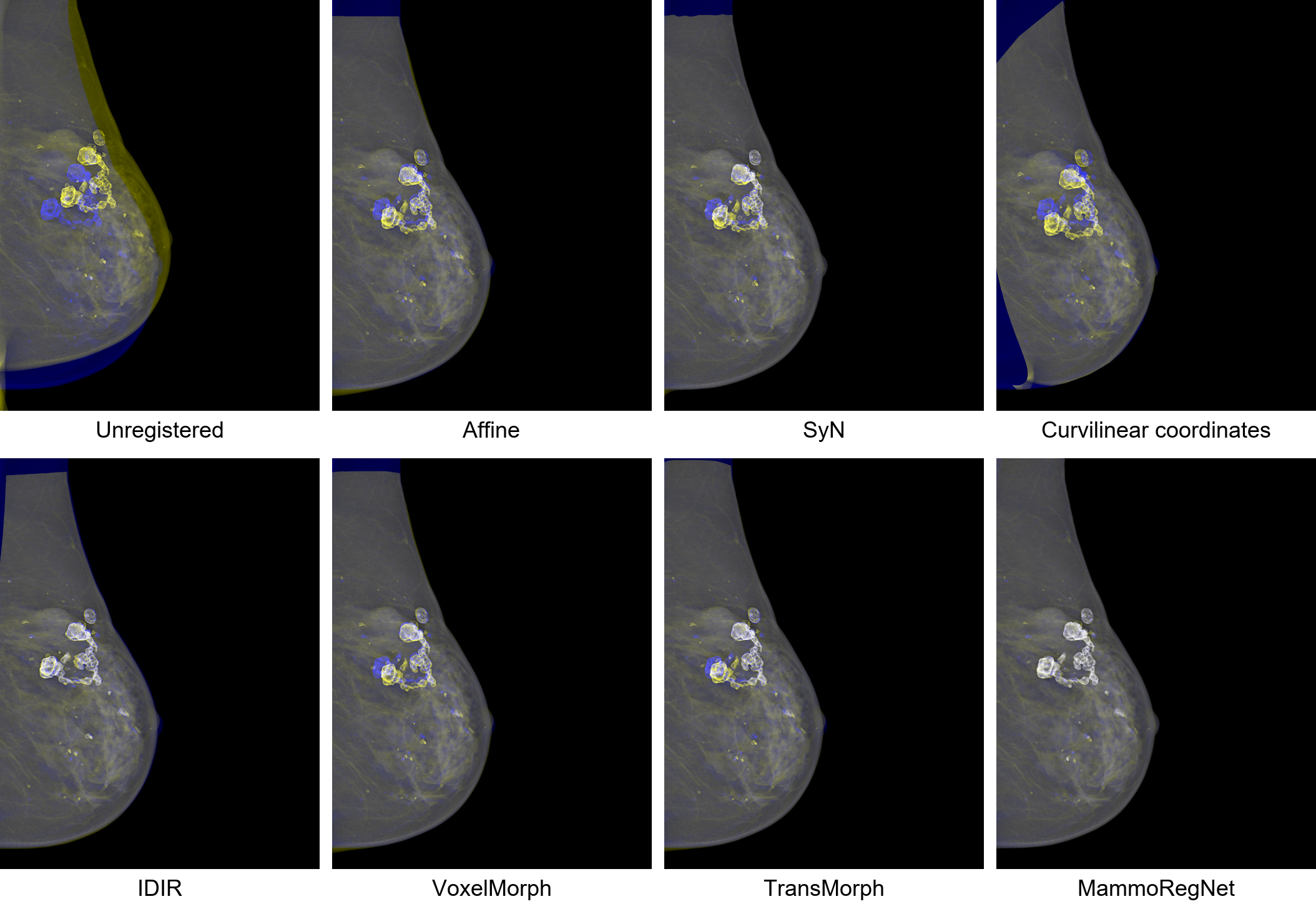}
    \end{minipage}
  }
  \caption{Registration results for an example pair in the L-CC and L-MLO projections. The overlays show the fixed image (in blue) and the moved image (in yellow) after registration using different methods: unregistered, Affine, SyN, Curvilinear Coordinates, IDIR, VoxelMorph, TransMorph, and MammoRegNet. Successful alignment is indicated by the overlap of corresponding anatomical structures in white.}
  \label{cc_ex}
\end{figure}

The quantitative evaluation of all registration methods on the MGRegBench benchmark is presented in Table~\ref{tab:results} and Table~\ref{tab:aux_results}, with visual examples of registration outcomes shown in Figure~\ref{cc_ex} for CC and MLO projections.
In addition to reporting the mean values for each metric, we provide the corresponding standard deviations (\textit{mean $\pm$ std}) to quantify the variability across evaluations and to highlight the notable instability exhibited by certain methods.

\begin{table*}[!ht]
\caption{Main quantitative evaluation of mammography registration methods on the MGRegBench dataset. The best results are in \textbf{bold}. $^{\dagger}$Training / Inference time or memory.}
\label{tab:results}
\resizebox{\linewidth}{!}{%
\begin{tabular}{@{}lcccc@{}}
\toprule
\textbf{Method} & \textbf{rTRE (\%) $\downarrow$} & \textbf{NJD (\%) $\downarrow$} & \textbf{VRAM (Gb) $\downarrow$} & \textbf{Runtime $\downarrow$}\\
\midrule
\textbf{Unreg.} & 4.5352 $\pm$ 2.5203 & -- & -- & --\\
\textbf{Affine} & 2.125991 $\pm$ 1.2868 & -- & -- & 2.1979 s\\
\textbf{SyN \citep{syn}} & 2.0869 $\pm$ 1.3116 & $3.23\cdot10^{-6}$ $\pm$ $3.21\cdot10^{-5}$ & -- & 8.3589 s\\
\textbf{Curv. coord.~\citep{cur}} & 4.0855 $\pm$ 4.1378 & 0.073 $\pm$ 0.213 & -- & 134.2862 s\\
\textbf{IDIR \citep{inr}} & 3.5754 $\pm$ 6.0633 & \textbf{0} & 8.5472 & 33.0461 s\\
\textbf{VM \citep{vxm}} & 2.125986 $\pm$ 1.2894 & $3.12\cdot10^{-4}\pm1.77\cdot10^{-3}$ & 4.076 / 0.43$^{\dagger}$ & \makecell{32.136 h / 0.349 s$^{\dagger}$}\\
\textbf{TM \citep{tm}} & 2.0995 $\pm$ 1.2928 & $1.9\cdot10^{-3}\pm1.53\cdot10^{-2}$ & 17.1727 / 1.815$^{\dagger}$ & \makecell{18.7336 h / 1.3908 s$^{\dagger}$}\\
\textbf{MRN \citep{thrun2025reconsidering}} & 2.2943 $\pm$ 1.6175 & 0.0863 $\pm$ 0.138 & 5.463 / \textbf{0.36}$^{\dagger}$ & \makecell{10.74 h / \textbf{0.2515 s}$^{\dagger}$}\\
\textbf{Affine +MRN \citep{thrun2025reconsidering}} & \textbf{1.9997 $\pm$ 1.2949} & 0.1115 $\pm$ 0.2225 & 5.463 / \textbf{0.36}$^{\dagger}$ & 10.95 h / 0.6678 s$^{\dagger}$\\
\bottomrule
\end{tabular}
}
\end{table*}

\begin{table*}[!ht]
\caption{Auxiliary intensity- and segmentation-based evaluation of mammography registration methods on the MGRegBench dataset. The best results are in \textbf{bold}.}
\label{tab:aux_results}
\resizebox{\linewidth}{!}{%
\begin{tabular}{@{}lccccc@{}}
\toprule
\textbf{Method} & \textbf{MSE $\downarrow$} & \textbf{SSIM $\uparrow$} & \textbf{MI $\uparrow$} & \textbf{CC $\uparrow$} & \textbf{DSC $\uparrow$}\\
\midrule
\textbf{Unreg.} & 782.48 $\pm$ 749.57 & 0.7587 $\pm$ 0.0845 & 1.1832 $\pm$ 0.0493 & 0.8261 $\pm$ 0.1243 & 0.9202 $\pm$ 0.0514\\
\textbf{Affine} & 452.26 $\pm$ 384.07 & 0.8025 $\pm$ 0.0833 & 1.2436 $\pm$ 0.051 & 0.8913 $\pm$ 0.0827 & 0.9702 $\pm$ 0.0226\\
\textbf{SyN \citep{syn}} & 379 $\pm$ 363.44 & 0.8211 $\pm$ 0.0796 & 1.2663 $\pm$ 0.0549 & 0.9113 $\pm$ 0.0755 & 0.977 $\pm$ 0.0205\\
\textbf{Curv. coord.~\citep{cur}} & 558.93 $\pm$ 468.37 & 0.7987 $\pm$ 0.0854 & 1.2356 $\pm$ 0.054 & 0.8647 $\pm$ 0.1064 & 0.9735 $\pm$ 0.0266\\
\textbf{IDIR \citep{inr}} & 595.12 $\pm$ 703.63 & 0.7942 $\pm$ 0.0928 & 1.2305 $\pm$ 0.0666 & 0.8635 $\pm$ 0.1585 & 0.9063 $\pm$ 0.1412\\
\textbf{VM \citep{vxm}} & 295.53 $\pm$ 249.06 & 0.8092 $\pm$ 0.0835 & 1.2611 $\pm$ 0.0501 & 0.9261 $\pm$ 0.0614 & 0.9806 $\pm$ 0.017\\
\textbf{TM \citep{tm}} & 227.74 $\pm$ 178.87 & 0.8139 $\pm$ 0.0824 & 1.2708 $\pm$ 0.049 & 0.941 $\pm$ 0.0507 & 0.9877 $\pm$ 0.0109\\
\textbf{MRN \citep{thrun2025reconsidering}} & 132.45 $\pm$ 86.02 & 0.8414 $\pm$ 0.0653 & 1.2985 $\pm$ 0.0418 & 0.9685 $\pm$ 0.0238 & 0.9922 $\pm$ 0.0066\\
\textbf{Affine +MRN \citep{thrun2025reconsidering}} & \textbf{130.65 $\pm$ 101.59} & \textbf{0.8455 $\pm$ 0.0663} & \textbf{1.3034 $\pm$ 0.0461} & \textbf{0.9688 $\pm$ 0.026} & \textbf{0.9923 $\pm$ 0.0067}\\
\bottomrule
\end{tabular}
}
\end{table*}

\subsection{Anatomical Accuracy and Registration Quality}

For anatomical accuracy, the combined Affine + MRN method achieves the best rTRE result (1.9997\%), followed closely by the classical ANTs (SyN) method (2.0869\%). Among deep learning approaches, VoxelMorph (2.125986\%) and TransMorph (2.0995 \%) demonstrate competitive landmark alignment. In contrast, standalone MammoRegNet (2.2943 $\pm$ 1.6175\%) and IDIR (3.5755 $\pm$ 6.0633\%) exhibit substantially higher mean errors, with particularly large standard deviations indicating significant sensitivity to outliers. The curvilinear coordinate method yields the worst landmark alignment (4.0855 $\pm$ 4.1378\%), confirming that its geometric parametrization is ill-suited for the complex, non-rigid deformations typical in mammography.

For intensity-based similarity metrics and segmentation-based overlap, the Affine + MRN pipeline consistently achieves the best scores, with the lowest MSE (130.65) and the highest SSIM (0.8455), MI (1.3034), CC (0.9688), and DSC (0.9923). Standalone MammoRegNet ranks a close second across these metrics. TransMorph, Voxelmorph follow, showing robust performance, particularly in MSE and CC. The curvilinear coordinates-based method shows moderate performance, as it matches the overall geometry rather than pixel-level intensities.

To summarize, the standalone MammoRegNet achieves strong performance in intensity-based similarity and segmentation overlap but exhibits a relatively high rTRE (2.2943\%). This degradation in landmark accuracy is driven by outliers where large initial misalignments lead to implausible deformations, highlighting the model’s sensitivity to global pose variations. In contrast, the Affine + MRN pipeline mitigates this issue for some cases by first resolving coarse misalignment through an affine transform, which significantly improves robustness. This hybrid approach not only reduces rTRE to the lowest among all methods (1.9997\%) but also further enhances intensity-based metrics. Representative failure cases for IDIR, Curvilinear coordinates method, and MammoRegNet are provided in the \ref{sec:outliers}.

\subsection{Deformation Regularity and Computational Efficiency}

The percentage of pixels with a negative Jacobian determinant (NJD) reflects the degree of folding in the deformation field. IDIR achieves a perfect NJD of 0\% while ANTs (SyN), VoxelMorph and TransMorph show a near-zero value ($3.23\cdot10^{-6}\%$, $3.12\cdot10^{-4}\%$, and $1.9\cdot10^{-3}\%$ respectively), indicating high smoothness. Among deep learning methods MammoRegNet (0.0863\%) and Affine + MRN (0.1115\%) exhibit substantially higher NJD, indicating noticeable folds in their deformation fields.

Finally, computational efficiency varies significantly. MammoRegNet offers the fastest inference (0.2515 s/pair) and the lowest VRAM usage during inference (0.36 GB) among the deep learning methods. The combined Affine + MRN pipeline has a longer inference time (0.6678 s) due to the sequential processing but retains the same memory footprint. VoxelMorph and TransMorph also provide fast inference but with higher memory consumption. In contrast, IDIR is slower (30 s/pair) and more memory-intensive (8.5 GB). ANTs runs in 8.36 s/pair on CPU, while the curvilinear method is the slowest.

\subsection{Qualitative Results and Failure Cases}

Visual inspection of the registration results reveals notable differences across methods. While IDIR does not achieve perfect alignment of the breast boundary, it demonstrates accurate correspondence of the internal masses, suggesting effective modeling of local deformations. In contrast, the curvilinear coordinates–based method yields the poorest visual alignment among all evaluated approaches. Moreover, due to its geometric formulation, this method truncates a portion of the breast tissue (it's particularly evident in the MLO projection) which may lead to loss of diagnostically relevant information. TransMorph exhibits slightly superior visual alignment compared to VoxelMorph (especially in the CC view) consistent with its more expressive transformer-based architecture. Nevertheless, ANTs, VoxelMorph, and TransMorph produce comparable overall performance: in all three cases, the masses are only partially overlapped, indicating residual misalignment. MammoRegNet with affine preprocessing achieves visually superior alignment compared to all other methods, with perfect overlap of both breast boundaries and internal anatomical structures.

IDIR is known to exhibit significant instability in practice. Figure~\ref{fig:idir} illustrates two representative cases where the method produces erratic deformations.

The registration method based on curvilinear coordinates is highly sensitive to the accurate selection of the landmark pair that defines the origin of the coordinate system in each image. Consequently,  errors in landmark placement lead to substantial registration failures (Figure~\ref{fig:cur_coords}).

MammoRegNet performs an affine alignment at a coarse, feature-level stage. Consequently, when images contain fine anatomical details, the method may produce suboptimal results. However, if a preliminary affine registration is applied directly at the image level before MammoRegNet, the overall performance improves significantly. For instance, Figure~\ref{fig:mrn1} demonstrates that such pre-alignment notably enhances registration quality.
Nevertheless, this strategy does not universally resolve all challenges: as shown in Figure~\ref{fig:mrn2}, there remain cases with severe outliers where both variants, original and pre-aligned, exhibit strong, unnatural deformations.

\section{Discussion}

\subsection{Principal Findings and Method Trade-Offs}

The trends observed in our benchmark largely mirror those reported in the broader medical image registration field. As discussed in~\citep{jena2024deep}, classical optimization-based methods such as ANTs (SyN)~\citep{syn} often remain competitive or superior to deep learning (DL) approaches when evaluated on anatomical accuracy. In our results, ANTs (SyN)~\citep{syn} achieves the second-best rTRE. However, the best anatomical accuracy is attained by a hybrid approach Affine + MRN~\citep{thrun2025reconsidering} that combines a classical affine step with a deep learning model. The SDM-MCs external validation supports the same high-level conclusion that MRN-based methods and strong general deep-learning baselines form the best-performing group, while also showing that exact rank order can be cohort-dependent. This suggests that integrating classical and learning-based paradigms can leverage their respective strengths, but should still be validated across datasets.

Our observations also align with findings from \citep{van2023robust}. While that INR-based approach demonstrated stable and smooth deformations across modalities, its evaluation on the DIRLab dataset \citep{castillo2009framework} showed that classical pTV algorithm \citep{vishnevskiy2016isotropic} still achieved superior accuracy compared to all deep learning methods, including implicit neural representations. This outcome is consistent with our finding that ANTs (SyN) remains highly competitive on the landmark-based rTRE metric.

\subsection{Clinical Relevance}

Accurate registration of longitudinal mammograms is a critical prerequisite for a range of clinical applications, including change detection, lesion tracking, and computer-aided diagnosis~\citep{lee2023enhancing, dadsetan2022deep, karaman2024longitudinal, wang2024ordinal}. Subtle temporal changes in breast tissue (e.g., the emergence or growth of masses, or microcalcification clusters) are often the earliest signs of malignancy~\citep{azam2021mammographic}. Reliable alignment of prior and current exams enables radiologists and AI systems to focus on true biological changes rather than misregistration artifacts caused by differences in patient positioning, compression, or anatomy. Recent work has shown that improving image-level alignment can also improve deformation-field quality and downstream breast cancer risk prediction, while jointly training alignment and risk prediction may introduce a trade-off between alignment quality and prediction performance~\citep{thrun2025reconsidering}. These findings underscore that registration quality is not merely a geometric objective, but a factor that quantitatively influences downstream clinical modeling, highlighting the need for standardized benchmarks such as MGRegBench to study these effects systematically and reproducibly.

The MGRegBench benchmark, with its expert-annotated anatomical landmarks and diverse clinical scenarios, including post-surgical cases with metallic clips and varying breast densities, provides a realistic testbed for evaluating registration algorithms under conditions that reflect routine screening practice. By establishing standardized protocols and performance metrics, our work supports the development of robust registration tools that can ultimately enhance diagnostic accuracy, reduce false positives in longitudinal screening, and facilitate quantitative monitoring of disease progression or treatment response.

\subsection{External Validation on SDM-MCs}

\begin{figure*}[!ht]
  \centering
  \includegraphics[width=0.95\linewidth]{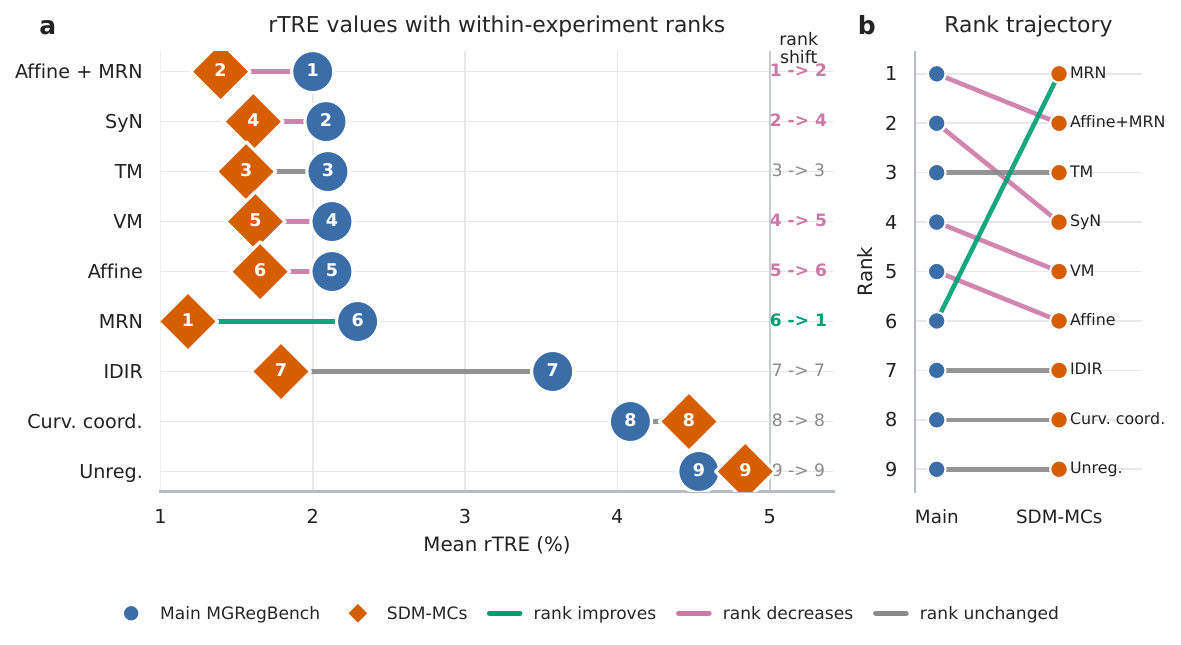}
  \caption{External-validation consistency of mean rTRE between MGRegBench and SDM-MCs. Methods are ordered by their main-benchmark rTRE rank. Marker numbers denote ranks within each experiment. Lower rTRE and lower rank are better. Line colors indicate whether a method's rank improves, decreases, or remains unchanged on SDM-MCs.}
  \label{fig:rtre_rank_consistency}
\end{figure*}

To assess generalizability beyond the MGRegBench benchmark, we evaluated the same baselines on the independent SDM-MCs longitudinal mammography dataset~\citep{loizidou2021breast}. We treat rTRE as the primary external-validation endpoint because it directly measures anatomical landmark accuracy; intensity- and mask-based metrics are reported as secondary evidence because they may be influenced by acquisition and contrast differences across datasets. The primary anatomical and deformation results are reported in Table~\ref{tab:results_source_wise}, auxiliary intensity- and mask-based metrics in Table~\ref{tab:external_aux_results}, and rTRE/rank consistency in Figure~\ref{fig:rtre_rank_consistency}.

\begin{table*}[!ht]
\caption{External validation on the SDM-MCs. The best rTRE and NJD results are in \textbf{bold}.}
\label{tab:results_source_wise}
\centering
\begin{tabular}{@{}lcc@{}}
\toprule
\textbf{Method} & \textbf{rTRE (\%) $\downarrow$} & \textbf{NJD (\%) $\downarrow$}\\
\midrule
\textbf{Unreg.} & 4.8403 $\pm$ 2.0878 & --\\
\textbf{Affine} & 1.6543 $\pm$ 1.0494 & --\\
\textbf{SyN \citep{syn}} & 1.61 $\pm$ 1.0926 & \textbf{0}\\
\textbf{Curv. coord.~\citep{cur}} & 4.4696 $\pm$ 4.6606 & 0.5432 $\pm$ 1.9418\\
\textbf{IDIR \citep{inr}} & 1.7914 $\pm$ 1.1369 & \textbf{0}\\
\textbf{VM \citep{vxm}} & 1.6235 $\pm$ 1.0301 & $3\cdot10^{-4}\pm1.02\cdot10^{-3}$\\
\textbf{TM \citep{tm}} & 1.5622 $\pm$ 0.9425 & $1.6\cdot10^{-3}\pm4.98\cdot10^{-3}$\\
\textbf{MRN \citep{thrun2025reconsidering}} & \textbf{1.1806 $\pm$ 0.901} & 0.1403 $\pm$ 0.1285\\
\textbf{Affine +MRN \citep{thrun2025reconsidering}} & 1.3943 $\pm$ 1.01 & 0.1271 $\pm$ 0.1325\\
\bottomrule
\end{tabular}
\end{table*}

\begin{table*}[!ht]
\caption{Auxiliary external-validation metrics on the SDM-MCs. The best results are in \textbf{bold}.}
\label{tab:external_aux_results}
\resizebox{\linewidth}{!}{%
\begin{tabular}{@{}lccccc@{}}
\toprule
\textbf{Method} & \textbf{MSE $\downarrow$} & \textbf{SSIM $\uparrow$} & \textbf{MI $\uparrow$} & \textbf{CC $\uparrow$} & \textbf{DSC $\uparrow$}\\
\midrule
\textbf{Unreg.} & 1016.91 $\pm$ 450 & 0.6948 $\pm$ 0.1017 & 1.1462 $\pm$ 0.0373 & 0.7446 $\pm$ 0.0706 & 0.8968 $\pm$ 0.0817\\
\textbf{Affine} & 902.41 $\pm$ 546.46 & 0.7325 $\pm$ 0.0988 & 1.1962 $\pm$ 0.0467 & 0.7858 $\pm$ 0.095 & 0.9571 $\pm$ 0.0408\\
\textbf{SyN \citep{syn}} & 788.37 $\pm$ 511.87 & 0.7523 $\pm$ 0.0949 & 1.2137 $\pm$ 0.0519 & 0.8151 $\pm$ 0.0955 & 0.9637 $\pm$ 0.0394\\
\textbf{Curv. coord.~\citep{cur}} & 1286.45 $\pm$ 1132.15 & 0.7226 $\pm$ 0.1113 & 1.1897 $\pm$ 0.0587 & 0.7221 $\pm$ 0.1448 & 0.9561 $\pm$ 0.0453\\
\textbf{IDIR \citep{inr}} & 734.83 $\pm$ 429.41 & 0.7421 $\pm$ 0.1028 & 1.2056 $\pm$ 0.0521 & 0.8248 $\pm$ 0.0656 & 0.9557 $\pm$ 0.0263\\
\textbf{VM \citep{vxm}} & 534.14 $\pm$ 351.25 & 0.7326 $\pm$ 0.1064 & 1.2215 $\pm$ 0.0529 & 0.868 $\pm$ 0.0847 & 0.9754 $\pm$ 0.0331\\
\textbf{TM \citep{tm}} & 404.45 $\pm$ 266.35 & 0.7392 $\pm$ 0.1062 & 1.2343 $\pm$ 0.0522 & 0.9022 $\pm$ 0.0494 & 0.9889 $\pm$ 0.0114\\
\textbf{MRN \citep{thrun2025reconsidering}} & 198.38 $\pm$ 110.14 & 0.7873 $\pm$ 0.0951 & 1.2735 $\pm$ 0.0538 & 0.9547 $\pm$ 0.021 & \textbf{0.9956 $\pm$ 0.0022}\\
\textbf{Affine +MRN \citep{thrun2025reconsidering}} & \textbf{191.88 $\pm$ 108.9} & \textbf{0.7887 $\pm$ 0.0958} & \textbf{1.2744 $\pm$ 0.054} & \textbf{0.9552 $\pm$ 0.0211} & 0.9952 $\pm$ 0.0028\\
\bottomrule
\end{tabular}
}
\end{table*}

The external validation results in Table~\ref{tab:results_source_wise} are positive overall: all registration methods reduce rTRE relative to the unregistered baseline, and the weakest cases remain the unregistered and curvilinear-coordinate results. The strongest group is also preserved, with MRN, Affine + MRN, TransMorph, SyN, VoxelMorph, and Affine all clustered below 1.7\% rTRE on SDM-MCs. However, the exact top ordering changes (Figure~\ref{fig:rtre_rank_consistency}): standalone MRN becomes the best rTRE method, while Affine + MRN remains second. This suggests that the benefit of external affine pre-alignment is cohort-dependent: it is valuable on MGRegBench, where the failure-case analysis indicates sensitivity to coarse pose variation and outliers, but may be less critical on SDM-MCs. The auxiliary metrics in Table~\ref{tab:external_aux_results} reinforce the same broad trend: MRN and Affine + MRN dominate intensity similarity and mask overlap, TransMorph is the strongest general deep-learning baseline, and SyN/IDIR retain the smoothest deformation fields.

\subsection{Reproducibility and Benchmark Utility}
MGRegBench targets a key gap in mammography registration research: the lack of a common public benchmark with anatomical ground truth. By combining (a) a large-scale train split, (b) an evaluation split with expert-annotated anatomical landmarks, and (c) standardized metrics and protocols, all supported by breast segmentation masks for every image in the dataset, the benchmark enables like-for-like comparison across classical, learning-based, and INR-based methods.

Importantly, we provide ready-to-run training and inference pipelines for all compared methods, implemented from scratch or adapted from the authors’ official implementations. This includes both mammography-specific approaches and widely used general-purpose medical image registration methods adapted to mammography. When no official implementation is available, we provide our own re-implementation to support reproducible baselines and fair comparisons.

\subsection{Ethics and Fairness Considerations}
MGRegBench is built from de-identified data released by the original dataset providers, and we follow their respective licenses and usage terms. We do not distribute any patient identifiers and only release derived annotations and code.

To mitigate fairness concerns in evaluation, we use stratified sampling for the evaluation set to balance standard views and approximately match age and density distributions between train and evaluation. Nonetheless, the benchmark may inherit demographic and acquisition biases from the source datasets.

\subsection{Scope and Future Work}

MGRegBench is designed as a controlled benchmark instance with a fixed, auditable, and reproducible evaluation protocol rather than as an exhaustive aggregation of all longitudinal mammography resources. This design enables like-for-like comparison of registration methods while limiting hidden variability in preprocessing, annotation policy, model selection, and evaluation. Although the SDM-MCs experiment provides independent external validation beyond MGRegBench, broader multi-institutional validation remains important to cover a wider range of scanners, populations, clinical settings, and acquisition protocols.

The current benchmark focuses on same-view, whole-image 2D mammography registration. Extensions to cross-view CC-MLO registration, lesion-centered registration, digital breast tomosynthesis, multimodal breast imaging, and downstream clinical tasks such as change detection, lesion tracking, risk prediction, and treatment-response assessment remain important directions for future work. In this sense, MGRegBench provides a reproducible technical foundation for studying how anatomical registration accuracy affects clinically meaningful longitudinal mammography analysis.

\section{Conclusions}

We have introduced MGRegBench, the first public benchmark dataset for mammography registration, comprising 100 expert-annotated image pairs and a large training set derived from INbreast, KAU-BCMD, and RSNA. By providing standardized data, metadata, and evaluation protocols our benchmark addresses critical gaps in reproducibility and method comparison that have long hindered progress in the field.

Among the evaluated baselines, Affine+MRN achieved the strongest overall performance under the proposed patient-disjoint MGRegBench protocol, albeit producing less smooth deformations and being prone to outliers. External validation on SDM-MCs preserved the main trend that MRN-based methods and strong general deep-learning baselines form the best-performing group, while showing cohort-dependent changes in exact rTRE ranking. To foster reproducible research and accelerate the development of registration methods, including methods for analyzing longitudinal disease progression, we publicly release our dataset and baseline implementations. We believe that the presented benchmark and dataset will serve as a vital resource for developing novel, specialized methods for mammography image registration and longitudinal disease progression analysis.





\bibliographystyle{elsarticle-num}
\bibliography{MGRegBench_arxiv}

\appendix

\section{Breast Segmentation Masks}

\begin{figure*}[!ht]
  \centering
  \subfloat[INBreast, R-MLO projection\label{dataset_with_masks-img-a}]{%
    \begin{minipage}{0.3\linewidth}
      \centering
        \includegraphics[width=1\linewidth]{imgs/inbreast.png}
    \end{minipage}
  }
  \hfill
  \subfloat[KAU-BCMD, R-CC projection\label{dataset_with_masks-img-b}]{%
    \begin{minipage}{0.3\linewidth}
      \centering
        \includegraphics[width=1\linewidth]{imgs/BC003261-4.png}
    \end{minipage}
  }
  \hfill
  \subfloat[RSNA, L-MLO projection\label{dataset_with_masks-img-c}]{%
    \begin{minipage}{0.3\linewidth}
      \centering
        \includegraphics[width=1\linewidth]{imgs/rsna.png}
    \end{minipage}
  }
  \\
  \subfloat[INBreast, R-MLO projection\label{dataset_with_masks-mask-a}]{%
    \begin{minipage}{0.3\linewidth}
      \centering
        \includegraphics[width=1\linewidth]{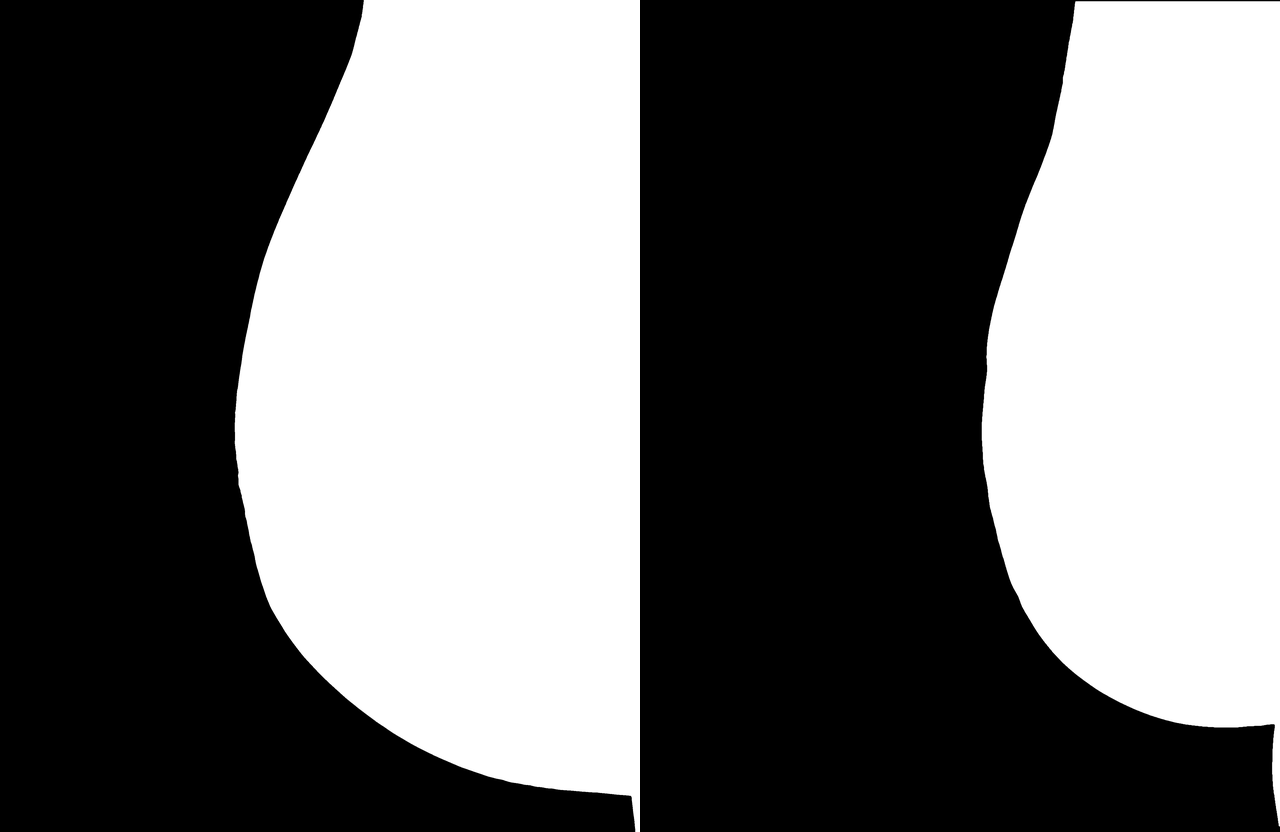}
    \end{minipage}
  }
  \hfill
  \subfloat[KAU-BCMD, R-CC projection\label{dataset_with_masks-mask-b}]{%
    \begin{minipage}{0.3\linewidth}
      \centering
        \includegraphics[width=1\linewidth]{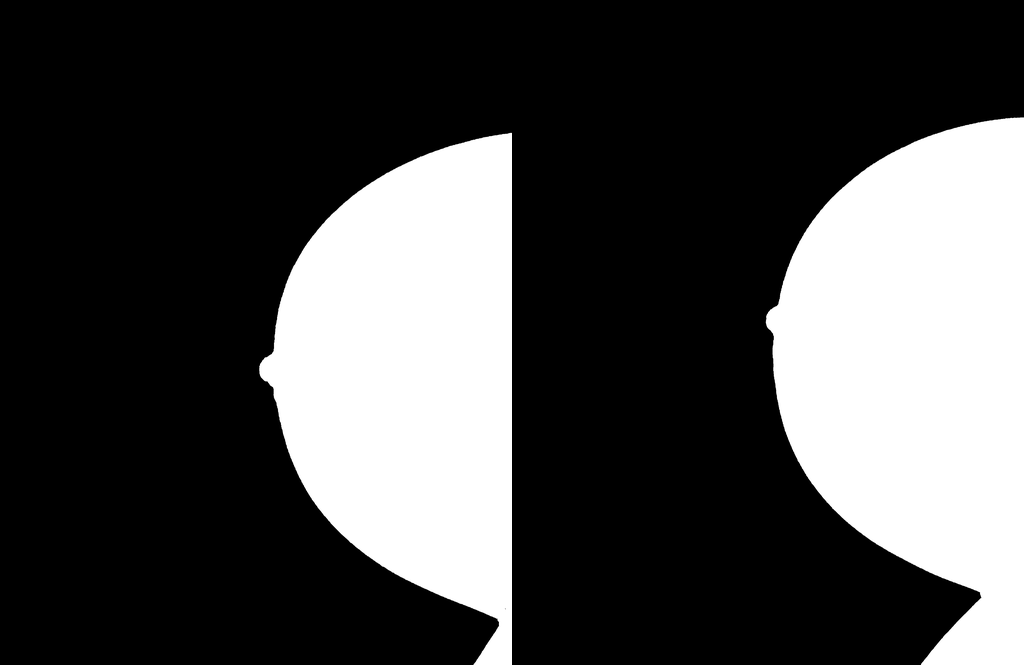}
    \end{minipage}
  }
  \hfill
  \subfloat[RSNA, L-MLO projection\label{dataset_with_masks-mask-c}]{%
    \begin{minipage}{0.3\linewidth}
      \centering
        \includegraphics[width=1\linewidth]{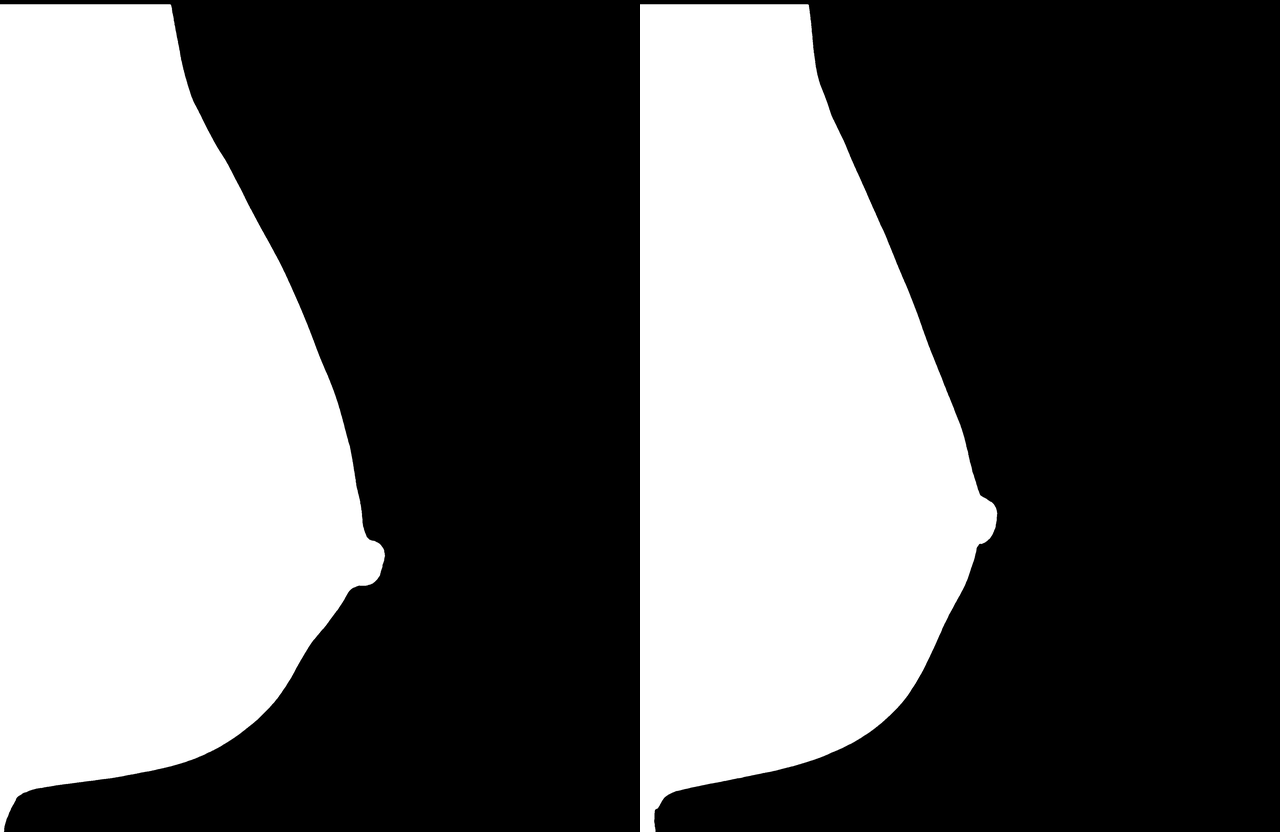}
    \end{minipage}
  }
  \caption{Registration-ready image pairs and corresponding breast segmentation masks from each source dataset in MGRegBench: (left) INBreast, (center) KAU-BCMD, and (right) RSNA. Top row shows original mammographic images; bottom row shows binary segmentation masks. Each pair represents the same breast in the same projection acquired during separate screening examinations.}
  \label{dataset_with_masks}
\end{figure*}

Figure~\ref{dataset_with_masks} shows registration-ready image pairs from each source dataset included in MGRegBench, along with their corresponding breast segmentation masks. The top row displays the original mammographic images, while the bottom row shows the binary segmentation masks.

\section{Failure Cases and Outlier Analysis}\label{sec:outliers}

\begin{figure*}[!ht]
  \centering
  \subfloat[Moving\label{moving_1_inr}]{%
    \begin{minipage}{0.3\linewidth}
      \centering
        \includegraphics[width=1\linewidth]{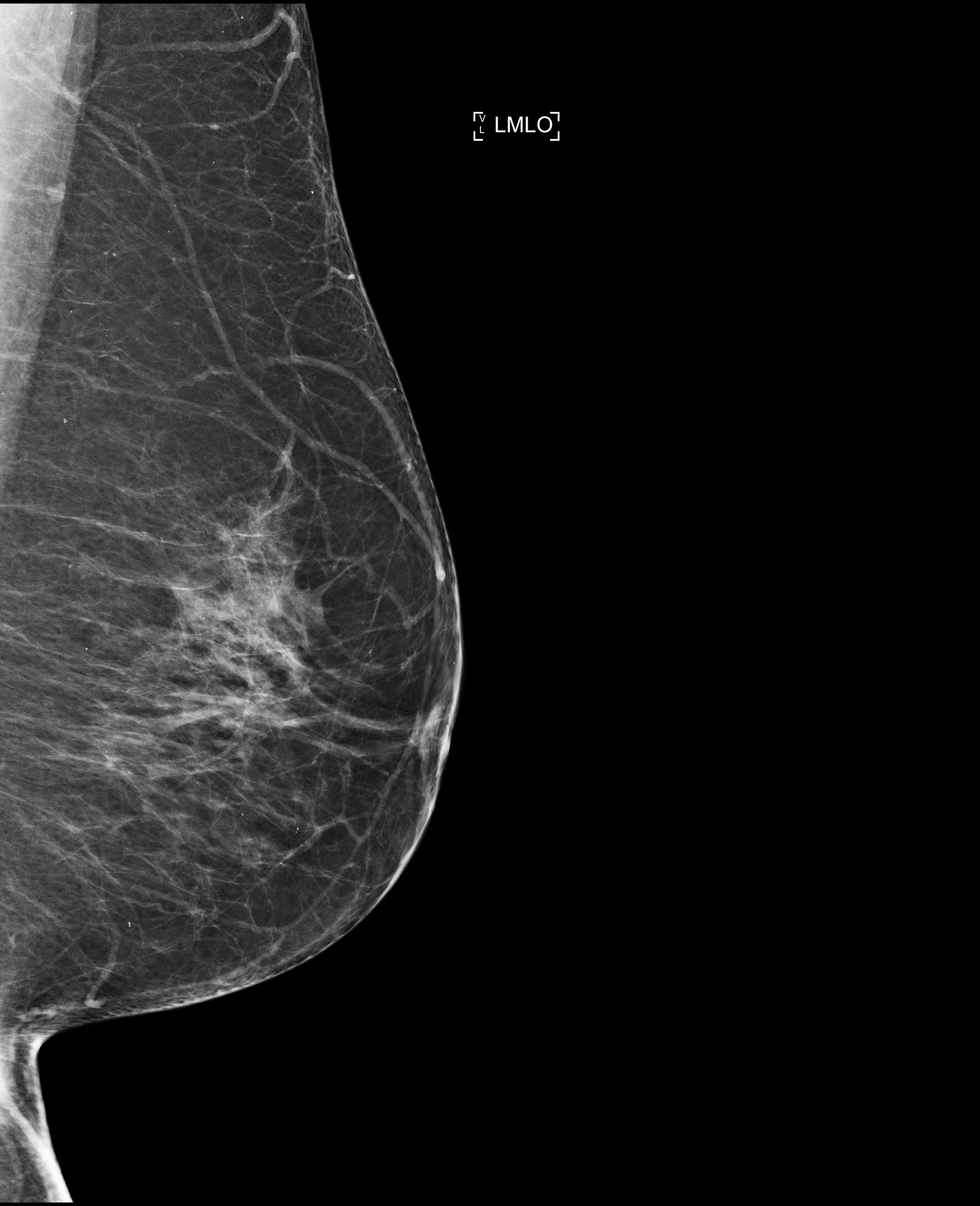}
    \end{minipage}
  }
  \hfill
  \subfloat[Fixed\label{fixed_1_inr}]{%
    \begin{minipage}{0.3\linewidth}
      \centering
        \includegraphics[width=1\linewidth]{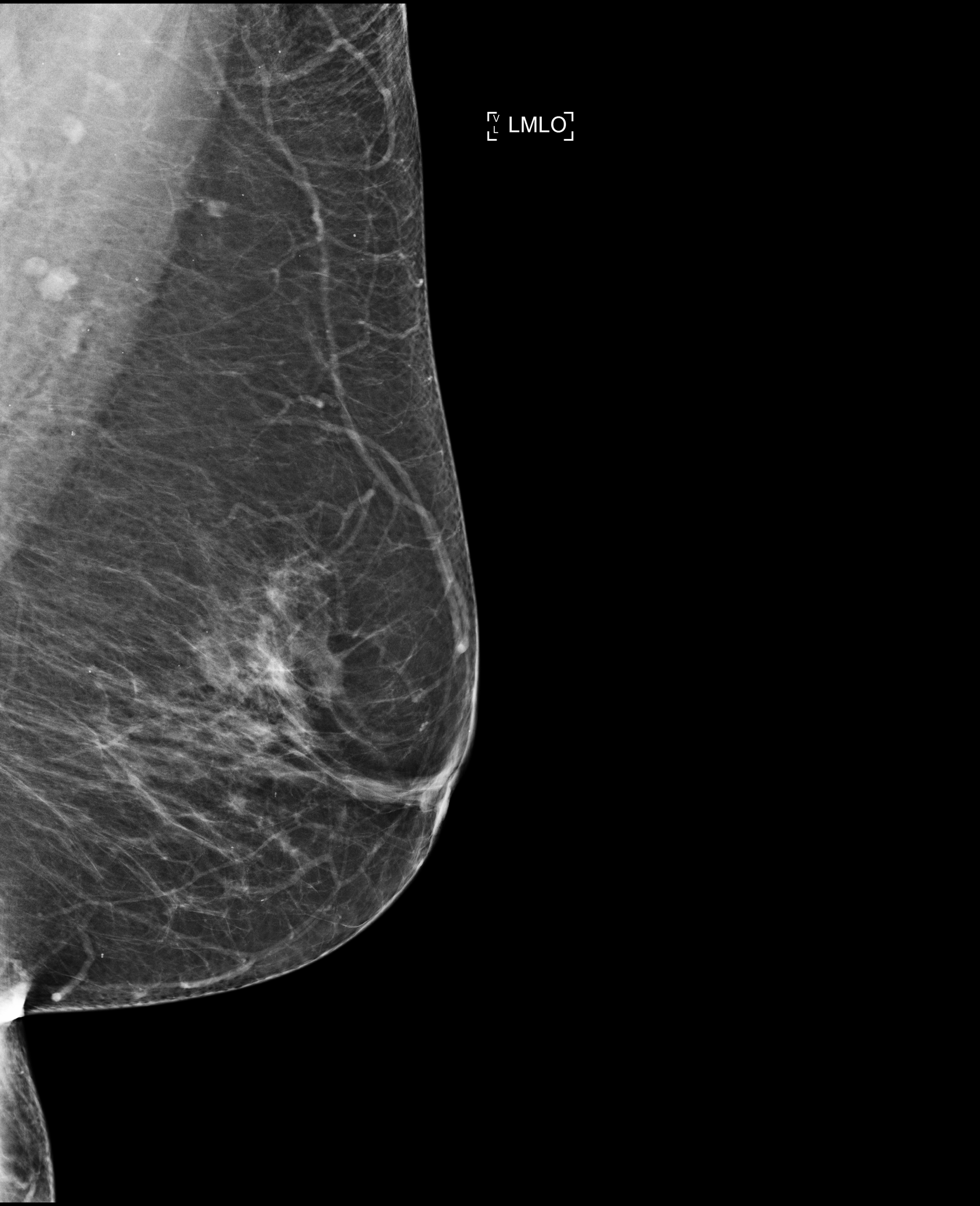}
    \end{minipage}
  }
  \hfill
  \subfloat[Moved\label{moved_1_inr}]{%
    \begin{minipage}{0.3\linewidth}
      \centering
        \includegraphics[width=1\linewidth]{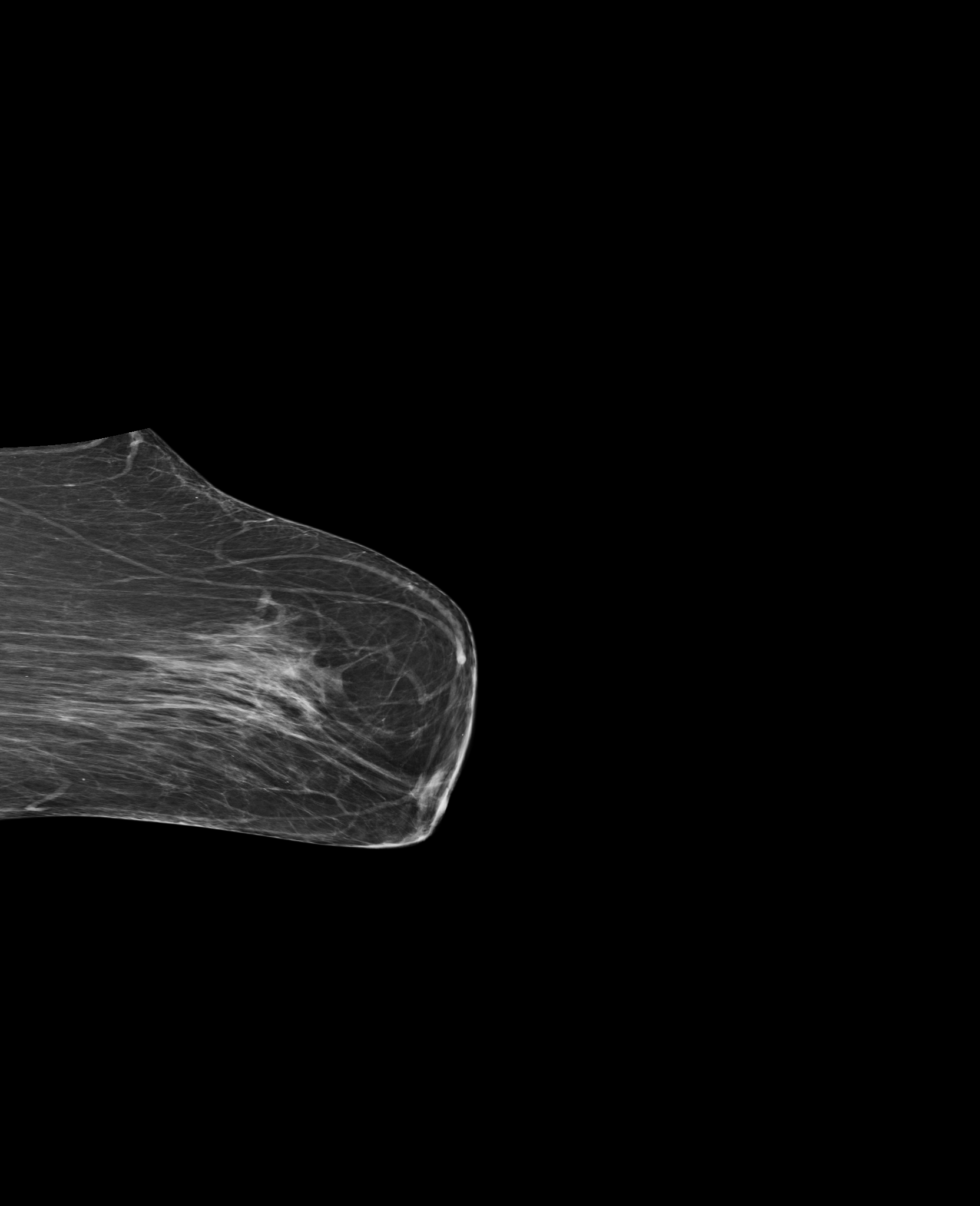}
    \end{minipage}
  }\\
  
  \subfloat[Moving\label{moving_2_inr}]{%
    \begin{minipage}{0.3\linewidth}
      \centering
        \includegraphics[width=1\linewidth]{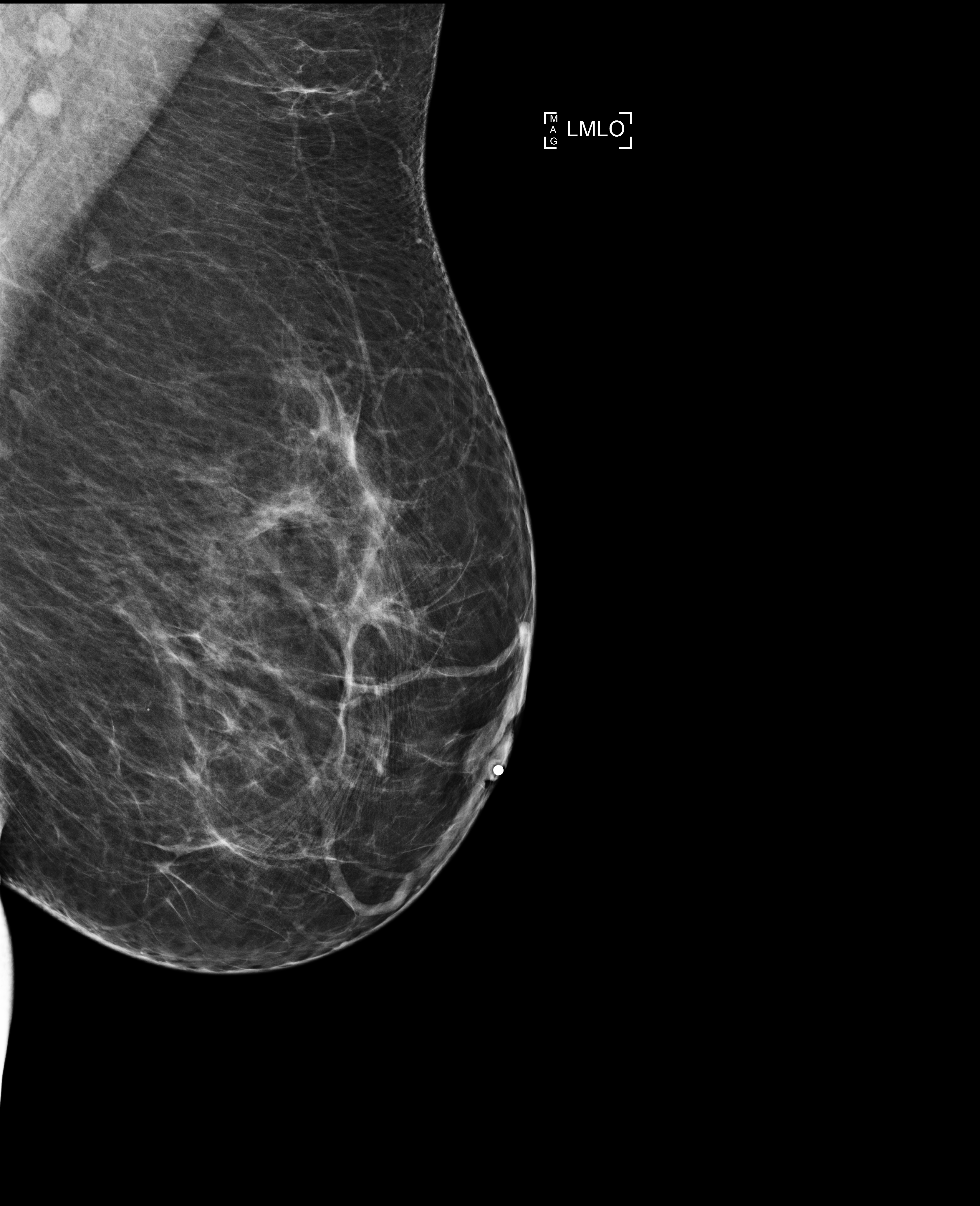}
    \end{minipage}
  }
  \hfill
  \subfloat[Fixed\label{fixed_2_inr}]{%
    \begin{minipage}{0.3\linewidth}
      \centering
        \includegraphics[width=1\linewidth]{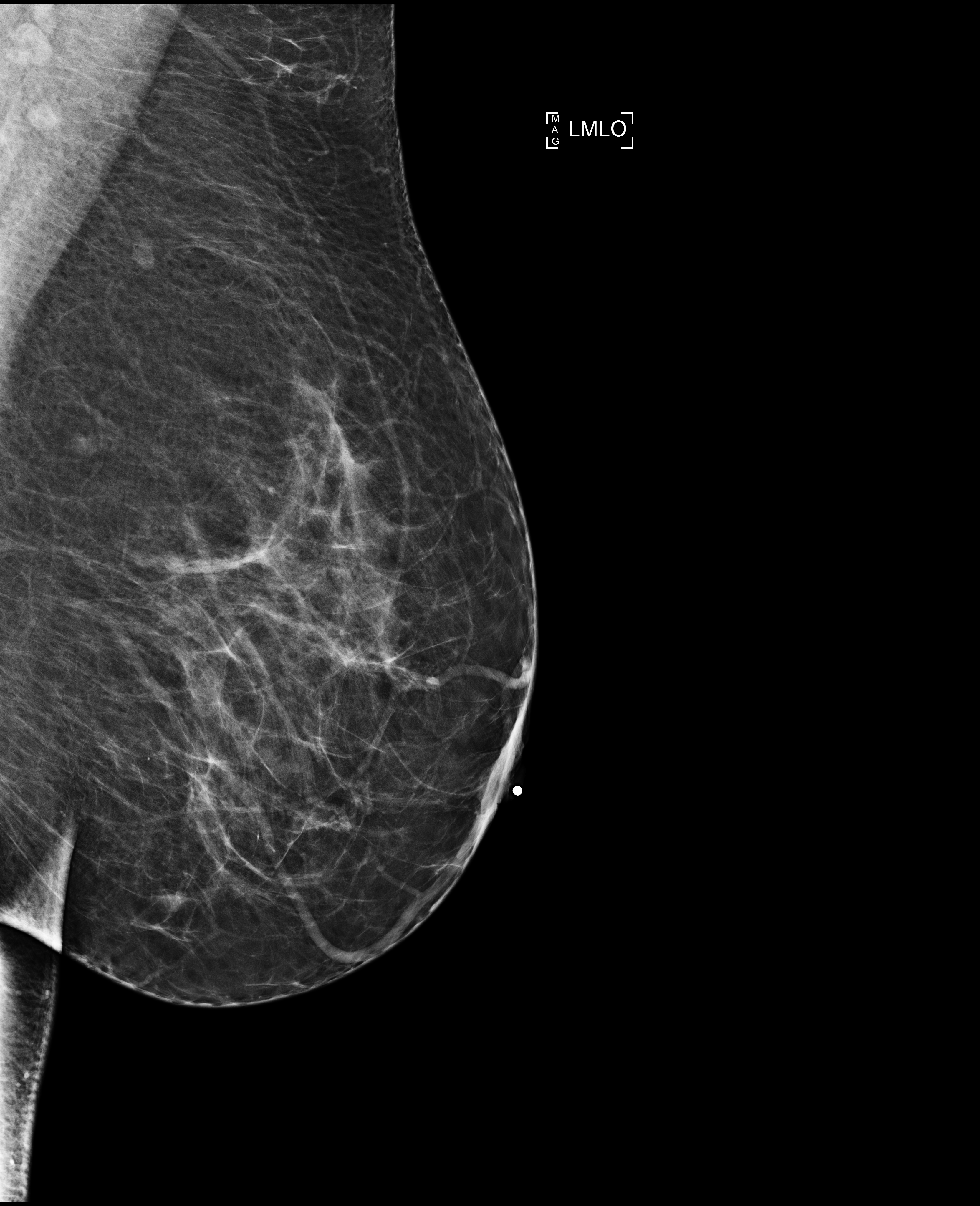}
    \end{minipage}
  }
  \hfill
  \subfloat[Moved\label{moved_2_inr}]{%
    \begin{minipage}{0.3\linewidth}
      \centering
        \includegraphics[width=1\linewidth]{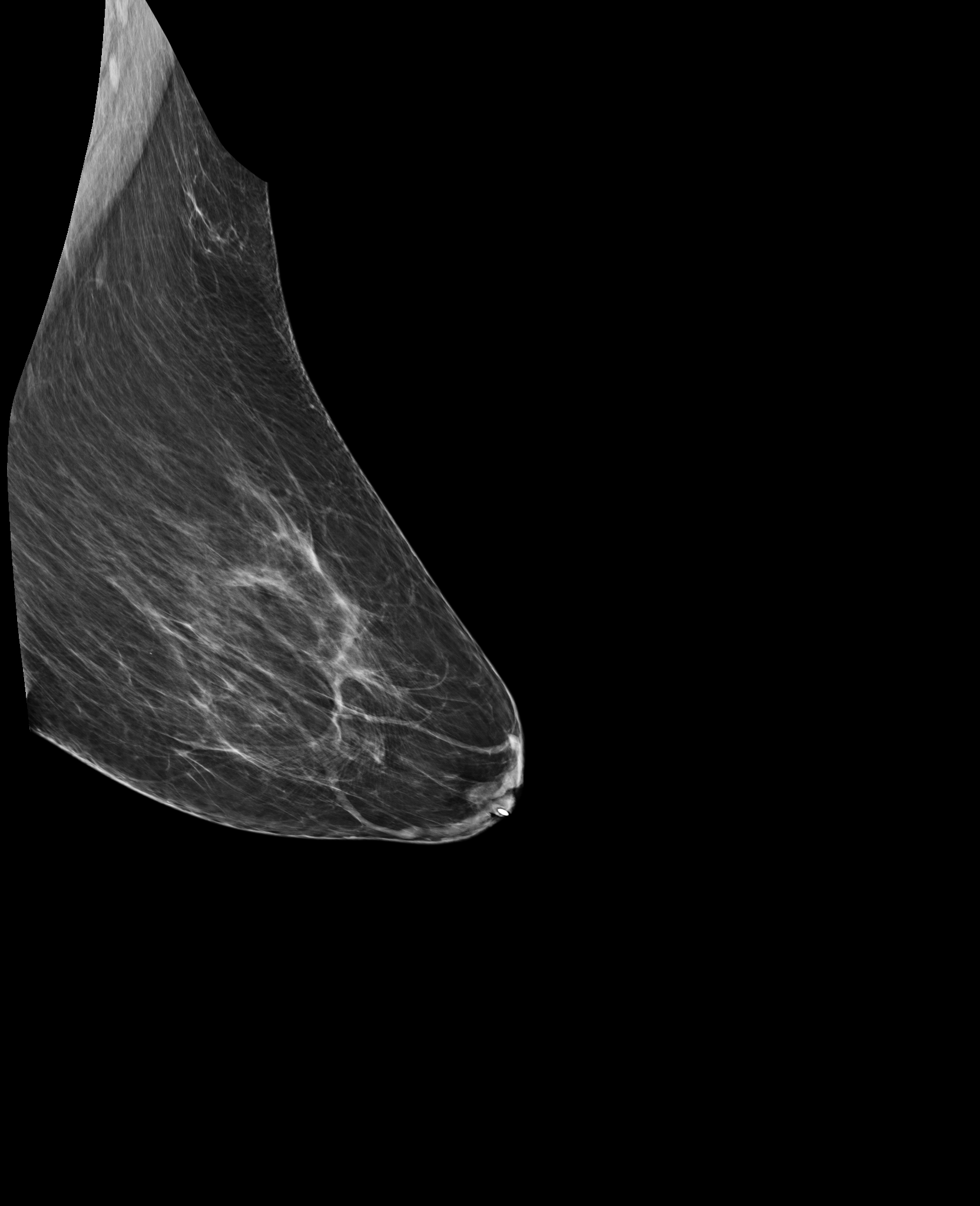}
    \end{minipage}
  }
  \caption{Qualitative examples of registration failures produced by IDIR on the MGRegBench dataset. In both cases, the method generates anatomically implausible deformations despite reasonable initial alignment, illustrating its instability.}
  \label{fig:idir}
\end{figure*}

\begin{figure*}[!ht]
  \centering
  \subfloat[Moving\label{moving_1_curv}]{%
    \begin{minipage}{0.3\linewidth}
      \centering
        \includegraphics[width=1\linewidth]{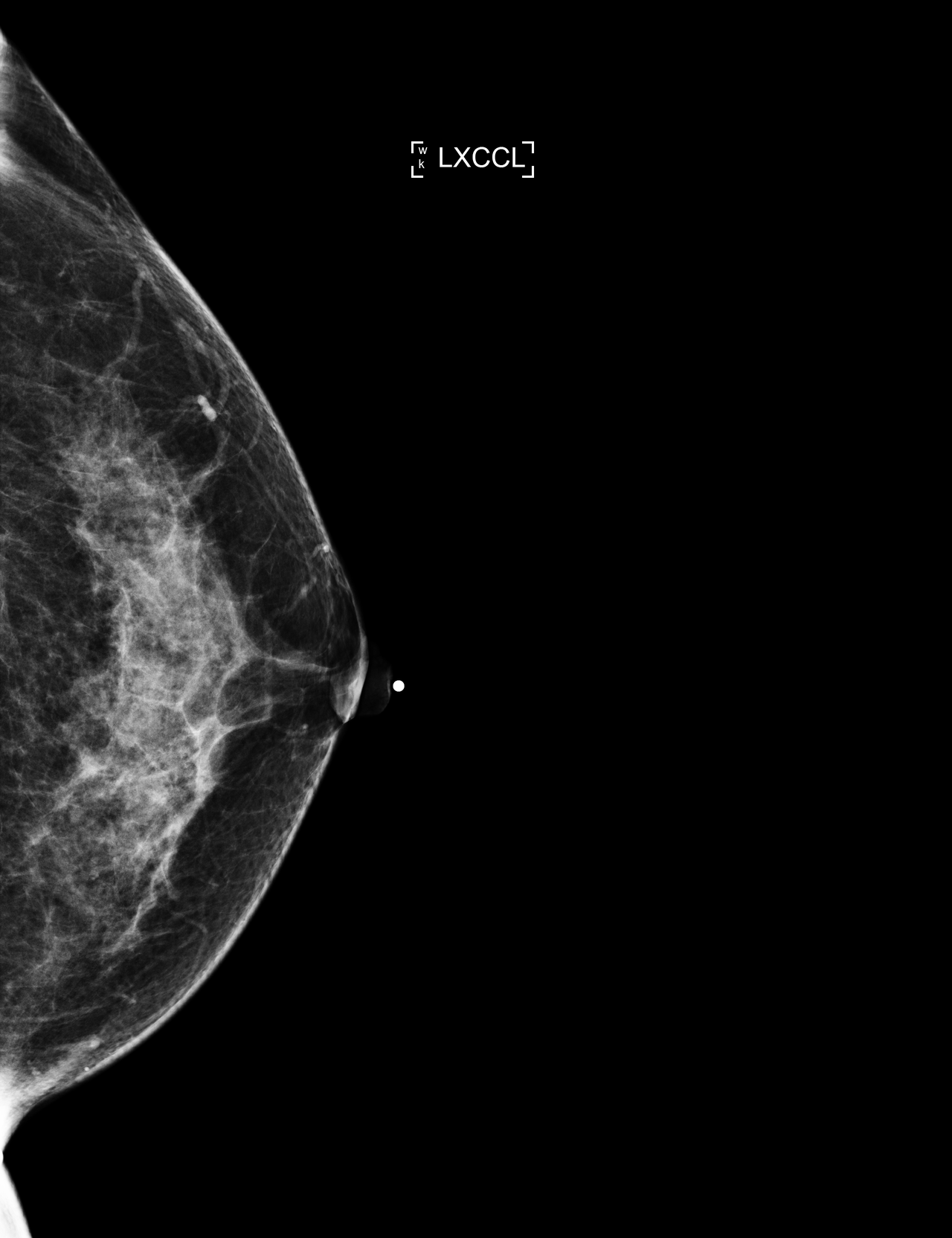}
    \end{minipage}
  }
  \hfill
  \subfloat[Fixed\label{fixed_1_curv}]{%
    \begin{minipage}{0.3\linewidth}
      \centering
        \includegraphics[width=1\linewidth]{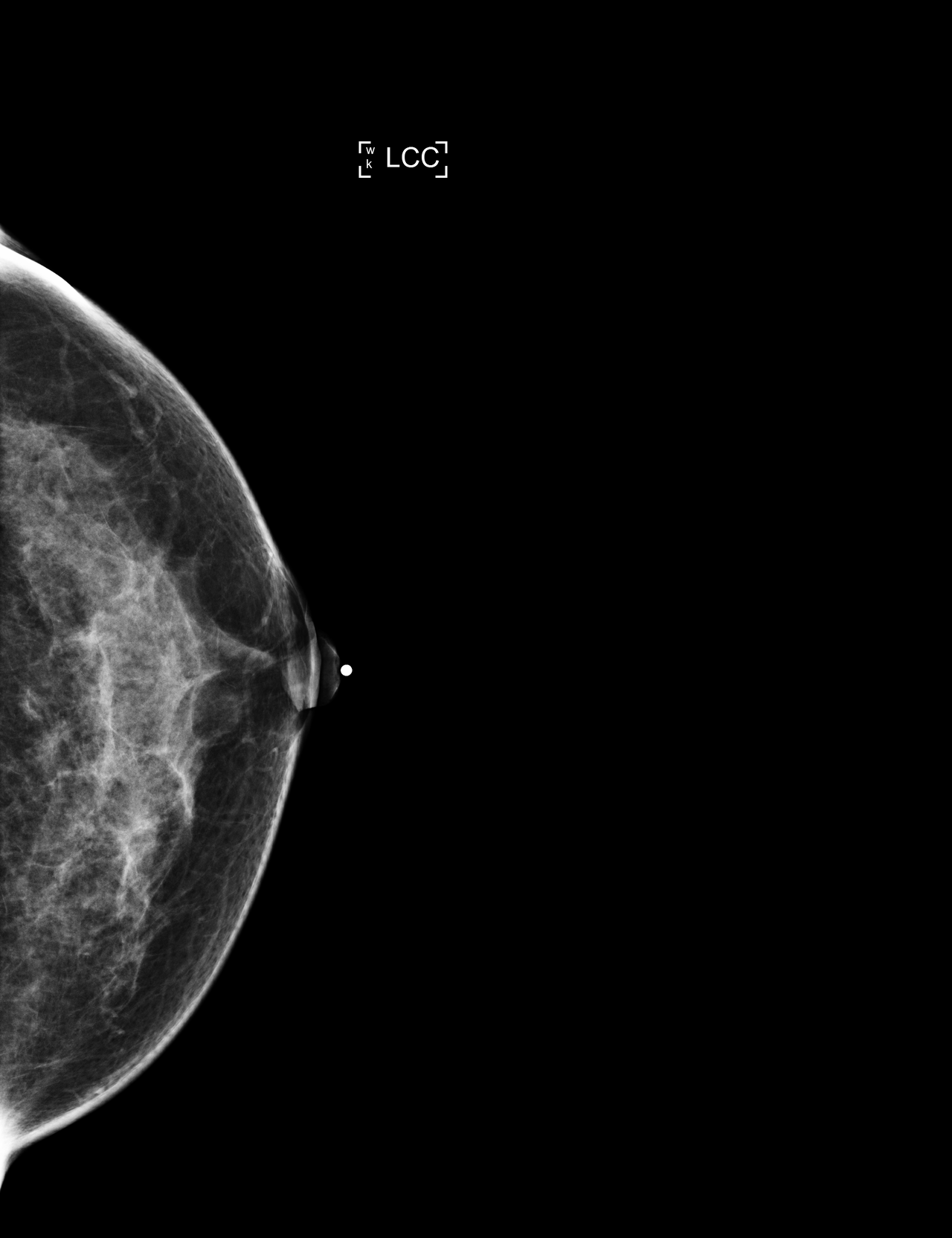}
    \end{minipage}
  }
  \hfill
  \subfloat[Moved\label{moved_1_curv}]{%
    \begin{minipage}{0.3\linewidth}
      \centering
        \includegraphics[width=1\linewidth]{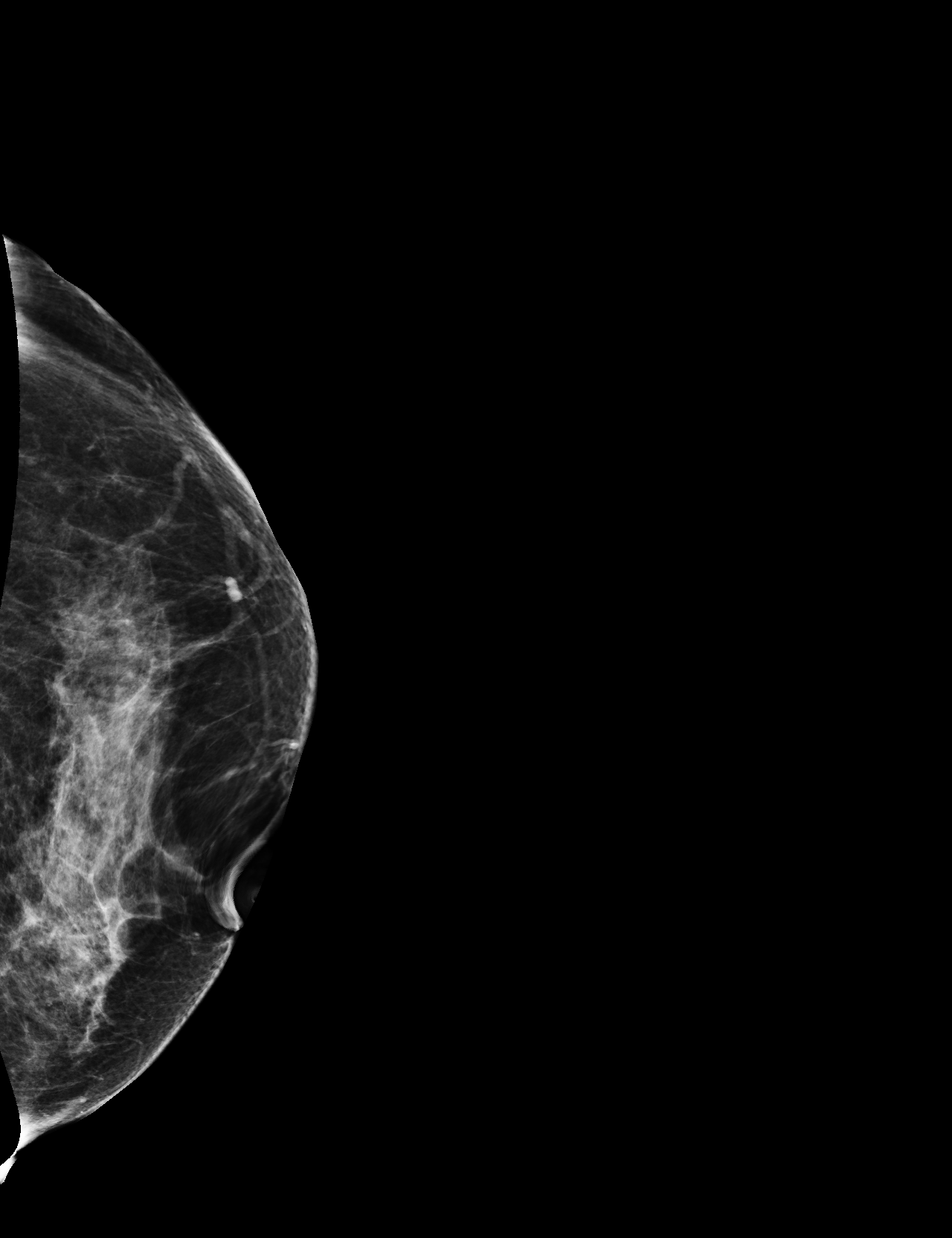}
    \end{minipage}
  }\\
  
  \subfloat[Moving\label{moving_2_curv}]{%
    \begin{minipage}{0.3\linewidth}
      \centering
        \includegraphics[width=1\linewidth]{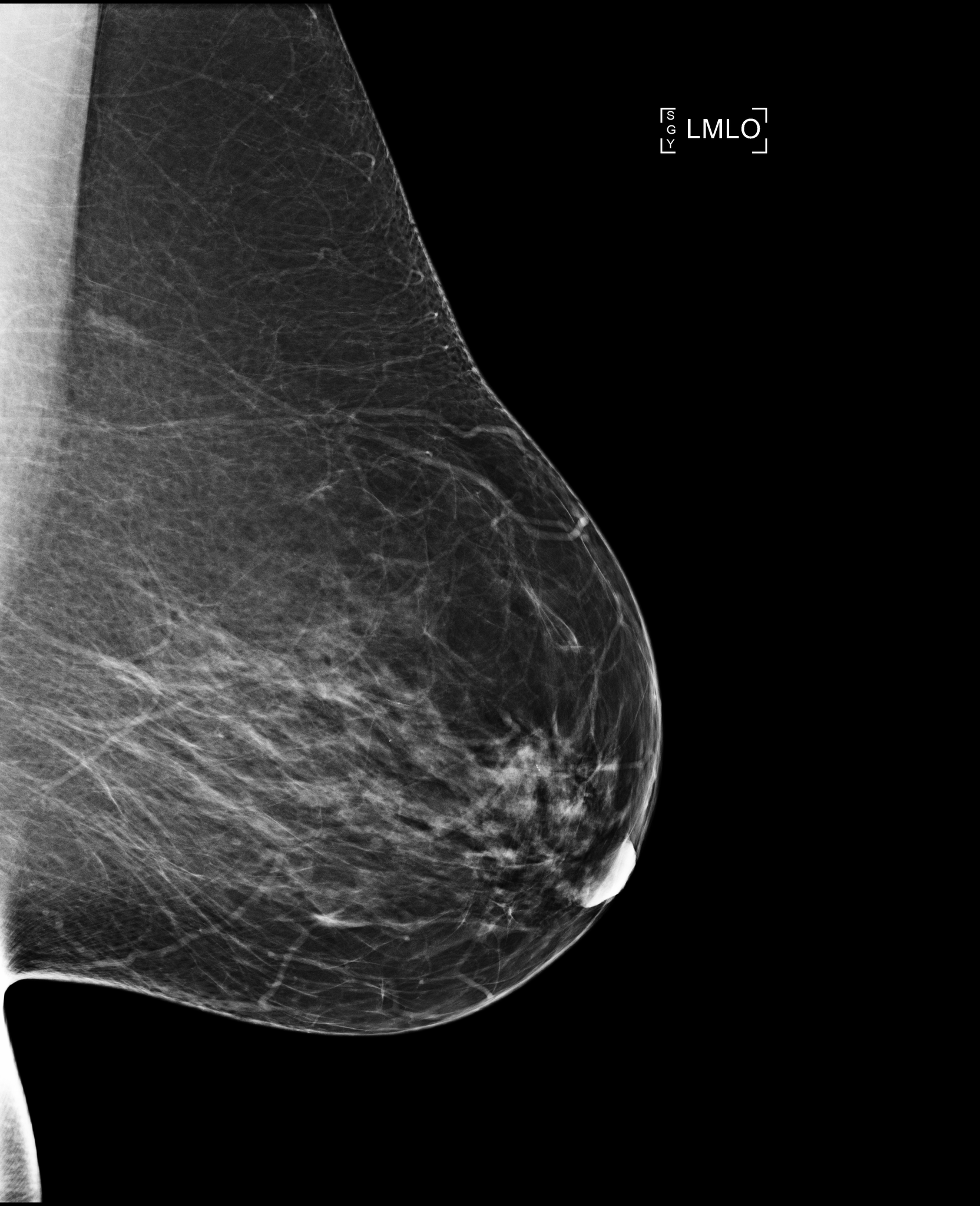}
    \end{minipage}
  }
  \hfill
  \subfloat[Fixed\label{fixed_2_curv}]{%
    \begin{minipage}{0.3\linewidth}
      \centering
        \includegraphics[width=1\linewidth]{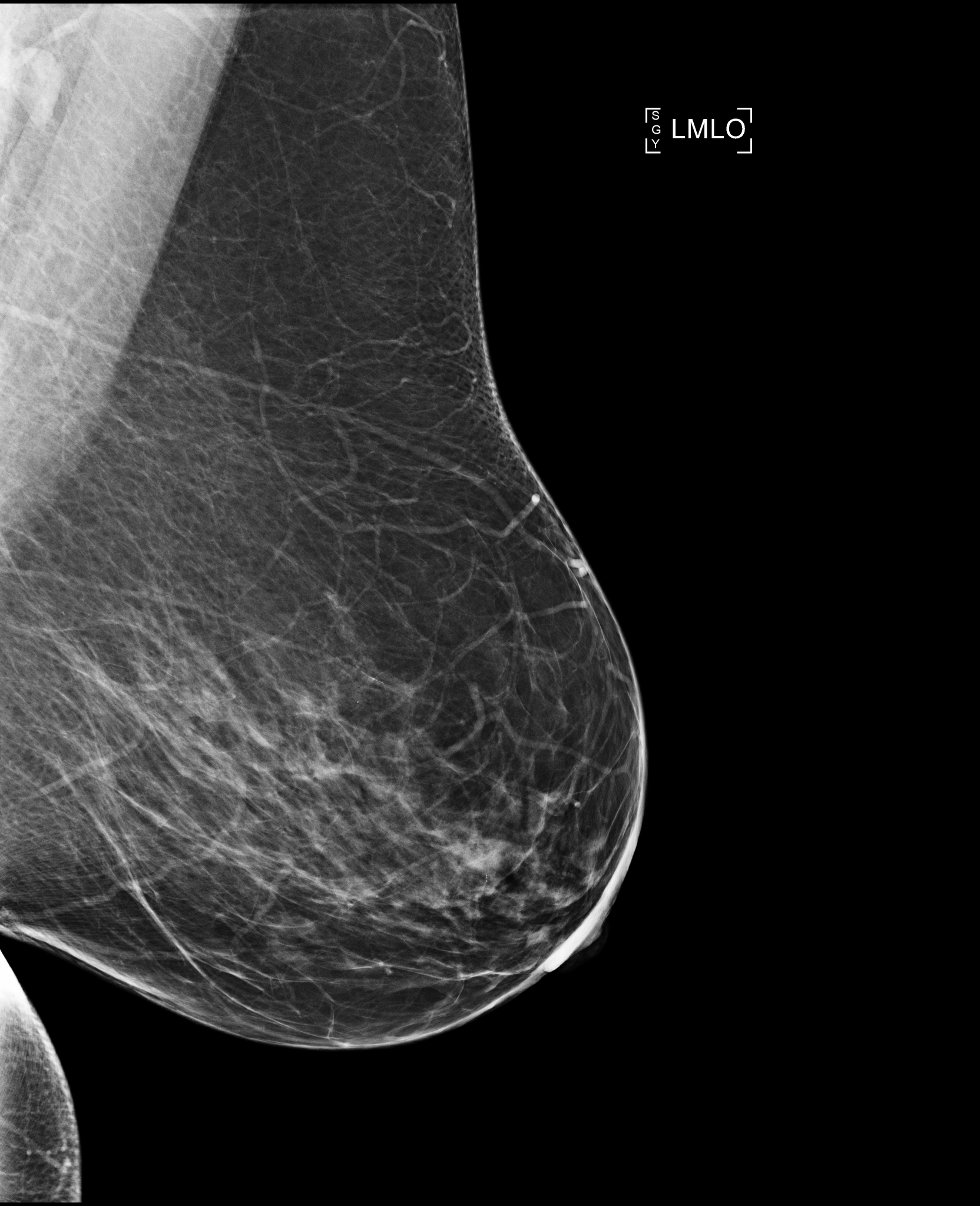}
    \end{minipage}
  }
  \hfill
  \subfloat[Moved\label{moved_2_curv}]{%
    \begin{minipage}{0.3\linewidth}
      \centering
        \includegraphics[width=1\linewidth]{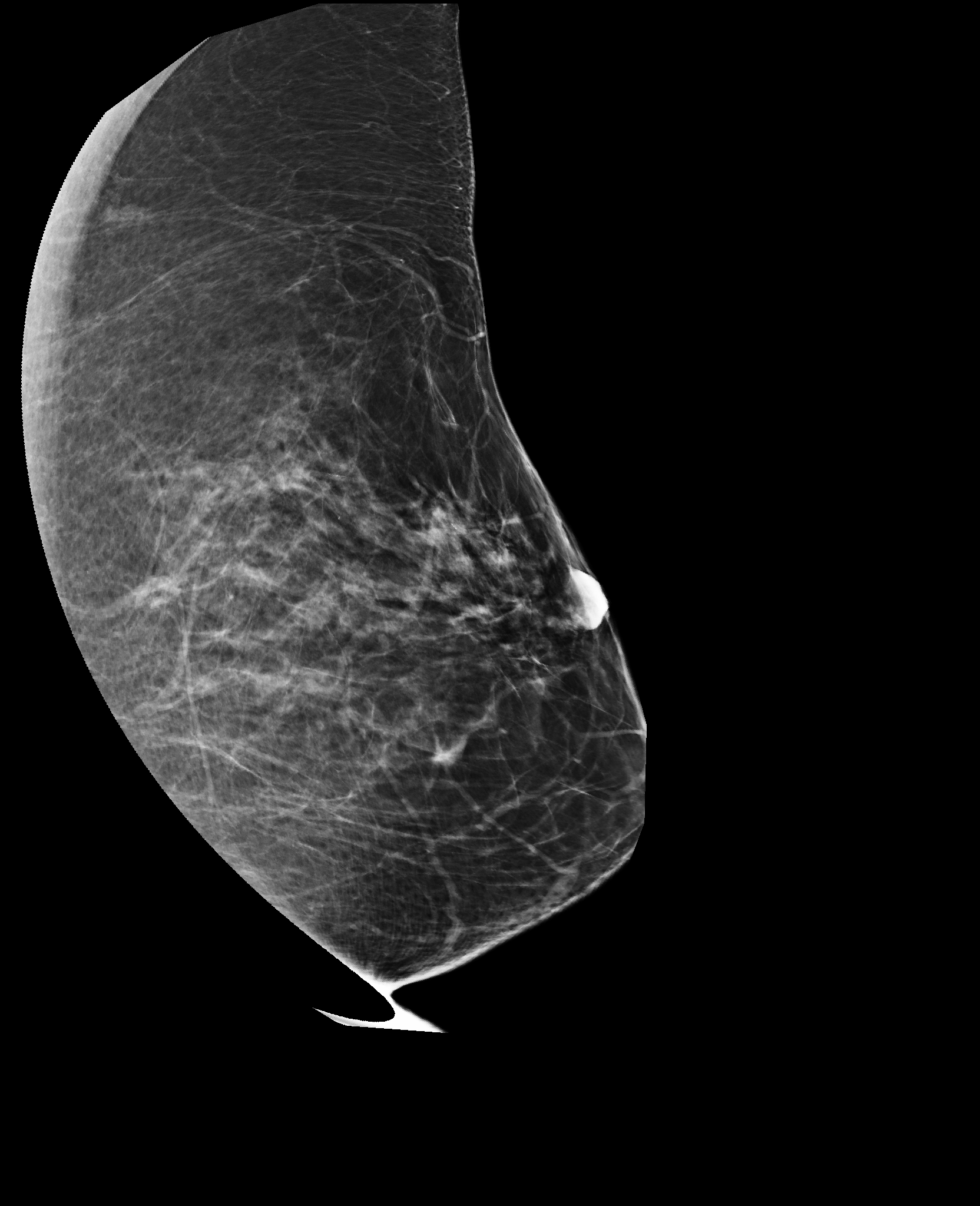}
    \end{minipage}
  }
  
  \caption{Failure cases of the curvilinear coordinate-based registration method on MGRegBench. Distortions occur due to incorrect selection of the origin points.}
  \label{fig:cur_coords}
\end{figure*}

\begin{figure*}[!ht]
  \centering
  \subfloat[Moving\label{a1}]{%
    \begin{minipage}{0.24\linewidth}
      \centering
        \includegraphics[width=1\linewidth]{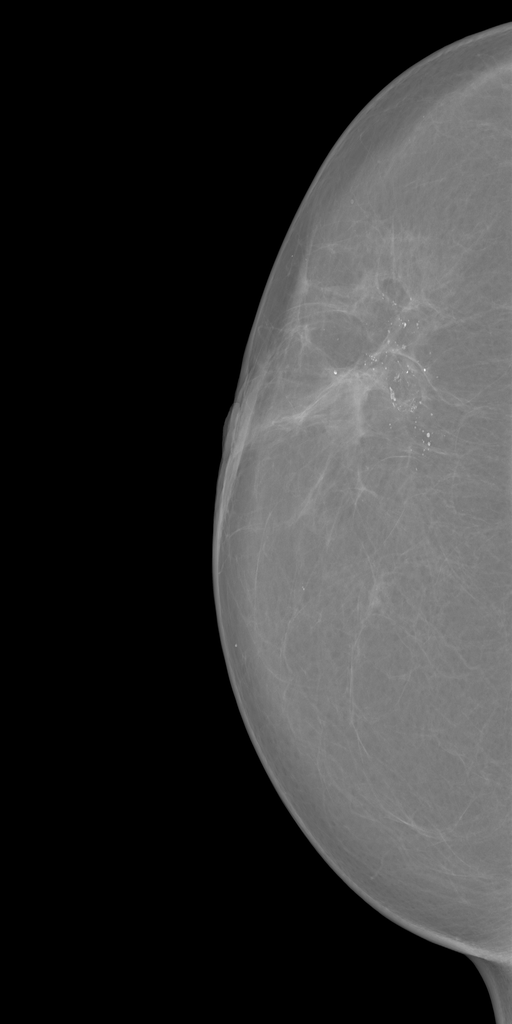}
    \end{minipage}
  }
  \hfill
  \subfloat[Fixed\label{b1}]{%
    \begin{minipage}{0.24\linewidth}
      \centering
        \includegraphics[width=1\linewidth]{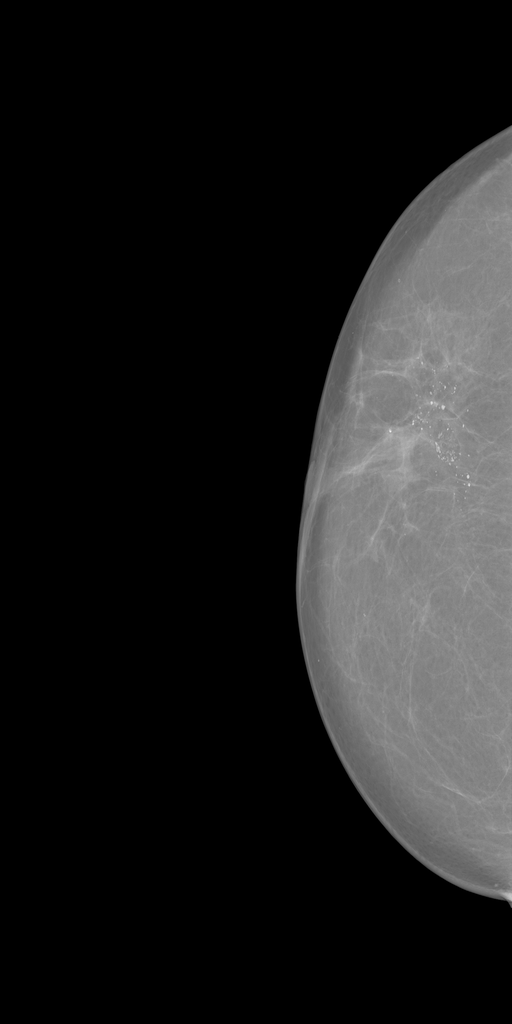}
    \end{minipage}
  }
  \hfill
  \subfloat[Standalone MRN result\label{c1}]{%
    \begin{minipage}{0.24\linewidth}
      \centering
        \includegraphics[width=1\linewidth]{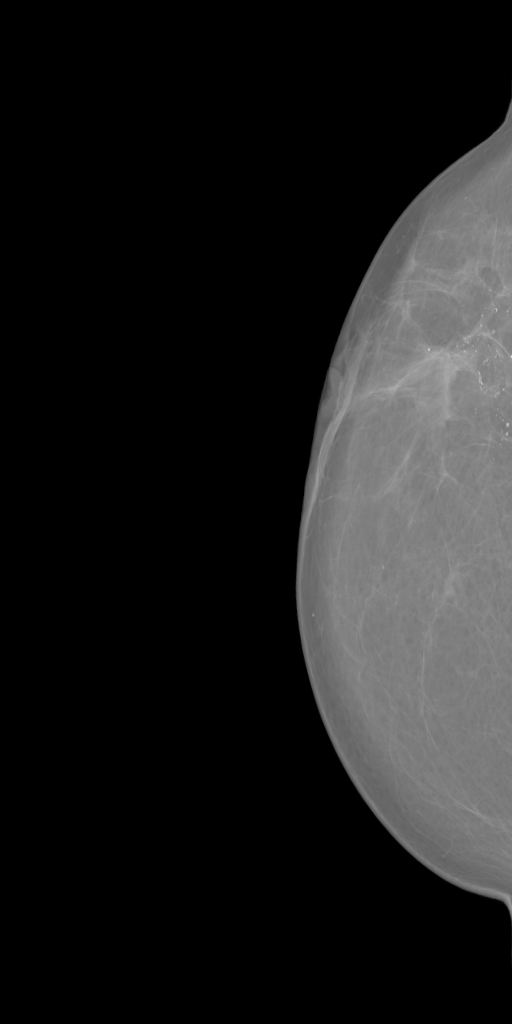}
    \end{minipage}
  }
  \hfill
  \subfloat[Affine + MRN result\label{d1}]{%
    \begin{minipage}{0.24\linewidth}
      \centering
        \includegraphics[width=1\linewidth]{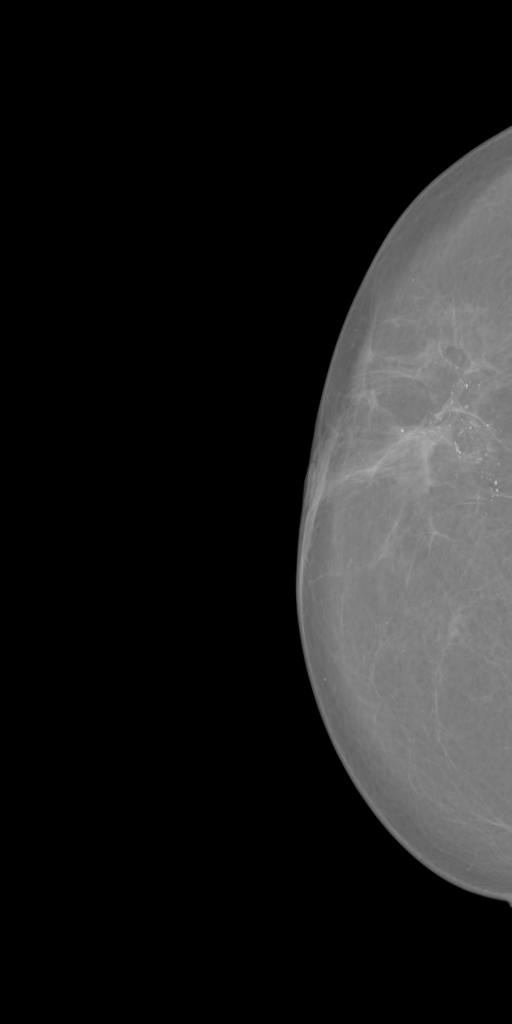}
    \end{minipage}
  }
  \bigskip
  \hfill
  \subfloat[Standalone MRN overlay\label{e1}]{%
    \begin{minipage}{0.24\linewidth}
      \centering
        \includegraphics[width=1\linewidth]{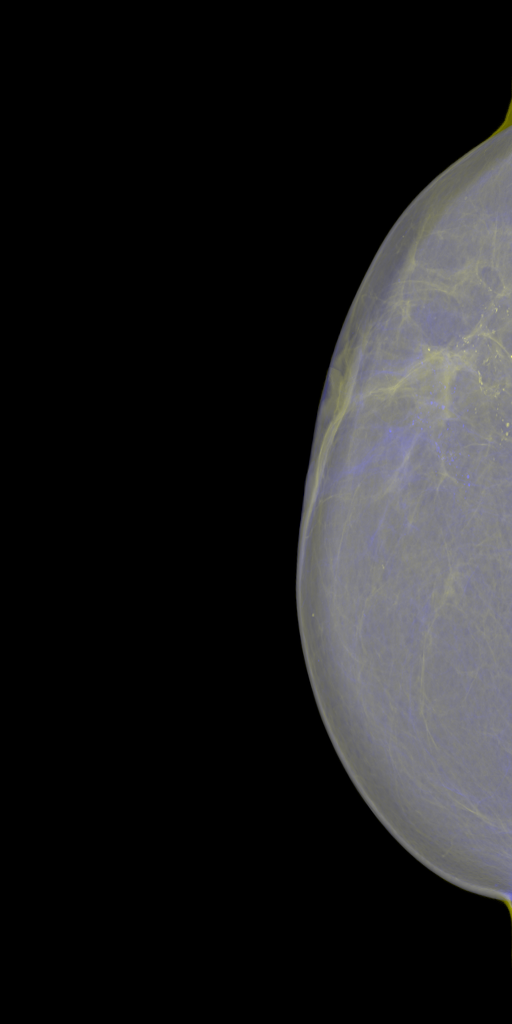}
    \end{minipage}
  }
  \hfill
  \subfloat[Affine + MRN overlay\label{f1}]{%
    \begin{minipage}{0.24\linewidth}
      \centering
        \includegraphics[width=1\linewidth]{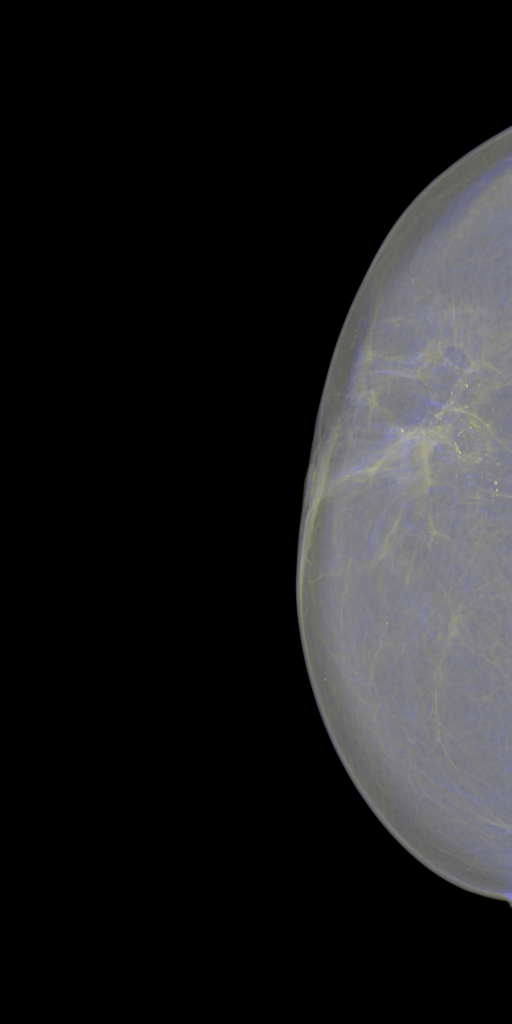}
    \end{minipage}
  }
  \hfill
  \hfill
  \caption{Qualitative comparison of MammoRegNet with and without affine pre-alignment on a representative mammogram pair. Top row: input moving and fixed images, followed by registration results using standalone MammoRegNet and MammoRegNet with affine pre-alignment. Bottom row: corresponding overlays (fixed in blue, registered moving in yellow). Affine pre-alignment significantly improves anatomical correspondence}
  \label{fig:mrn1}
\end{figure*}

\begin{figure*}[!ht]
  \centering
  \subfloat[Moving\label{a2}]{%
    \begin{minipage}{0.24\linewidth}
      \centering
        \includegraphics[width=1\linewidth]{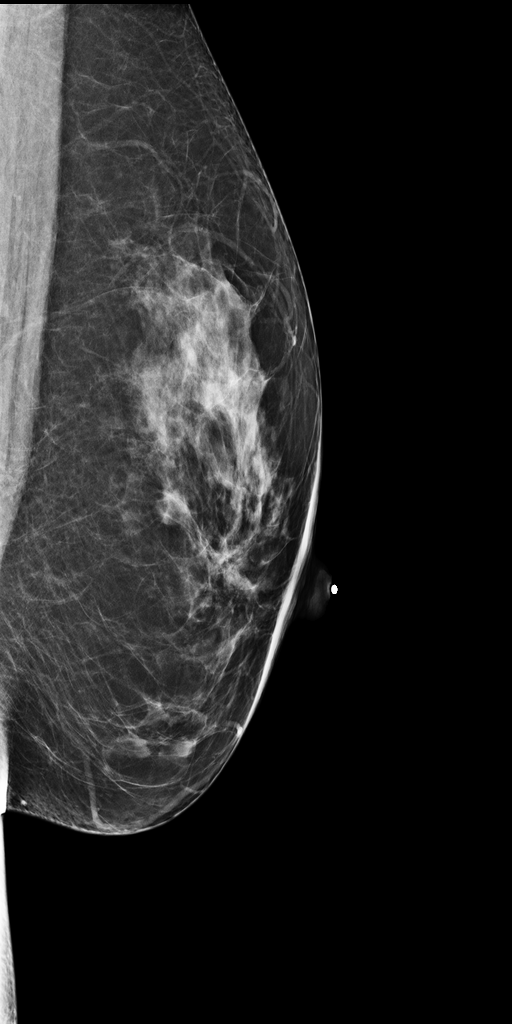}
    \end{minipage}
  }
  \hfill
  \subfloat[Fixed\label{b2}]{%
    \begin{minipage}{0.24\linewidth}
      \centering
        \includegraphics[width=1\linewidth]{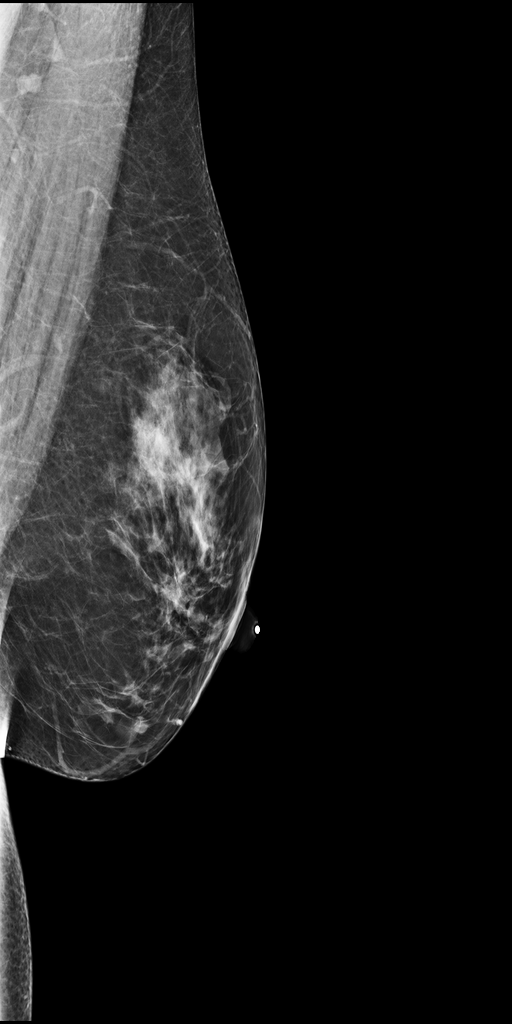}
    \end{minipage}
  }
  \hfill
  \subfloat[Standalone MRN result\label{c2}]{%
    \begin{minipage}{0.24\linewidth}
      \centering
        \includegraphics[width=1\linewidth]{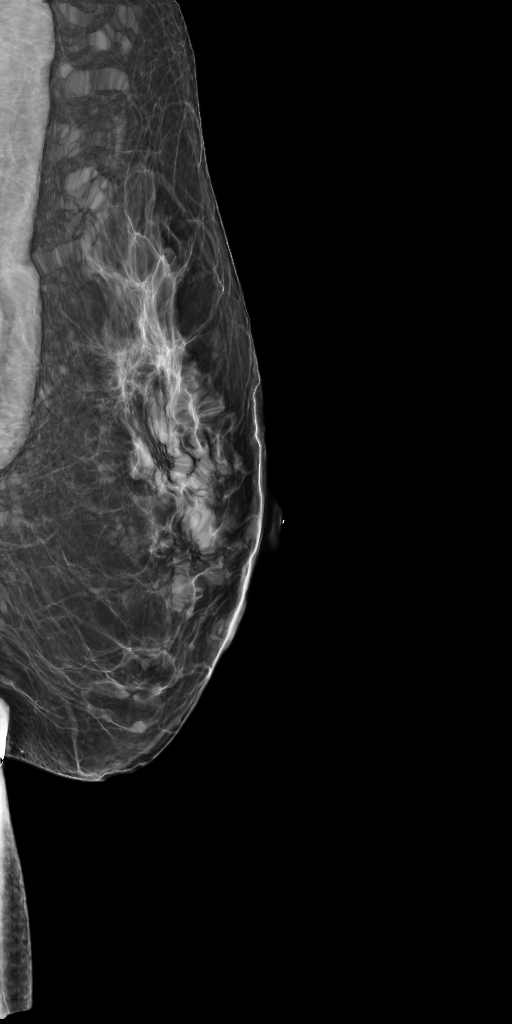}
    \end{minipage}
  }
  \hfill
  \subfloat[Affine + MRN result\label{d2}]{%
    \begin{minipage}{0.24\linewidth}
      \centering
        \includegraphics[width=1\linewidth]{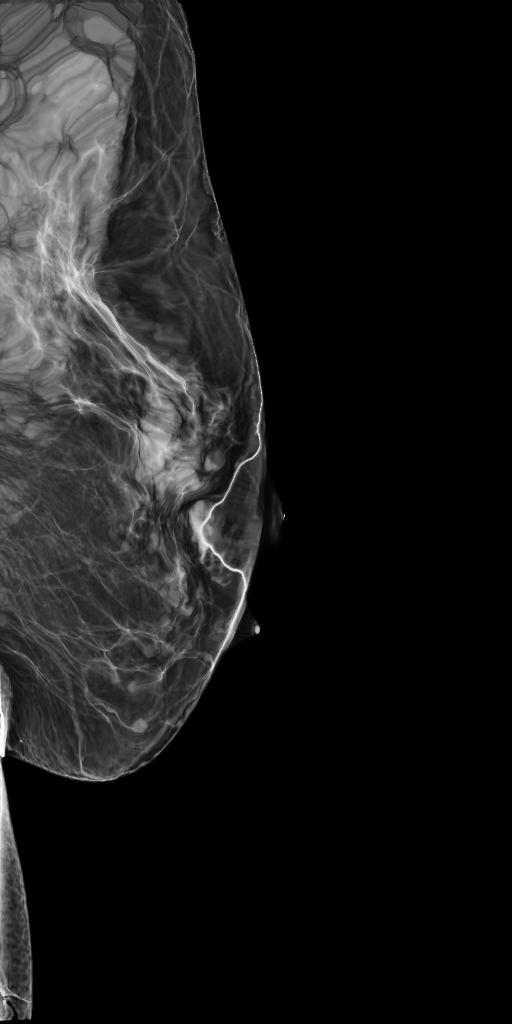}
    \end{minipage}
  }
  \bigskip
  \hfill  
  \subfloat[Standalone MRN overlay\label{e2}]{%
    \begin{minipage}{0.24\linewidth}
      \centering
        \includegraphics[width=1\linewidth]{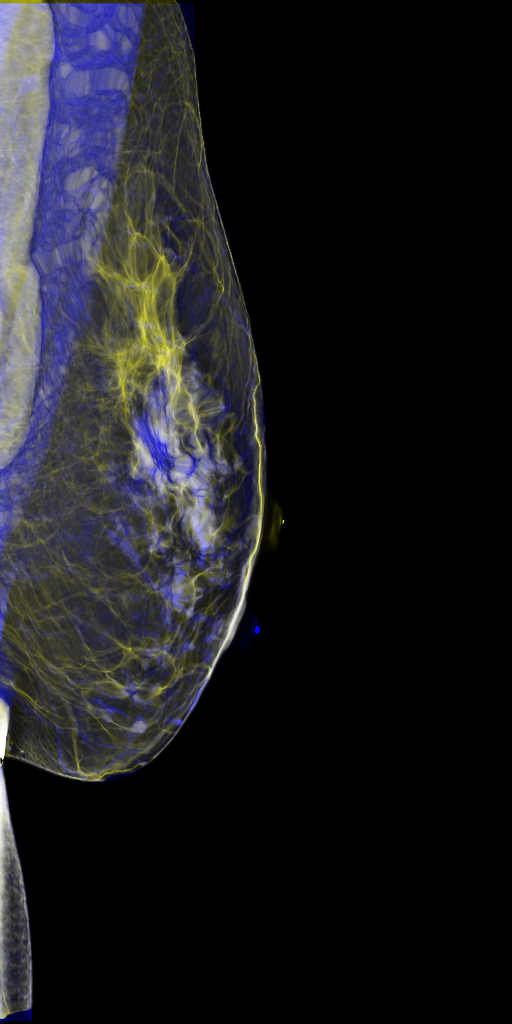}
    \end{minipage}
  }
  \hfill
  \subfloat[Affine + MRN overlay\label{f2}]{%
    \begin{minipage}{0.24\linewidth}
      \centering
        \includegraphics[width=1\linewidth]{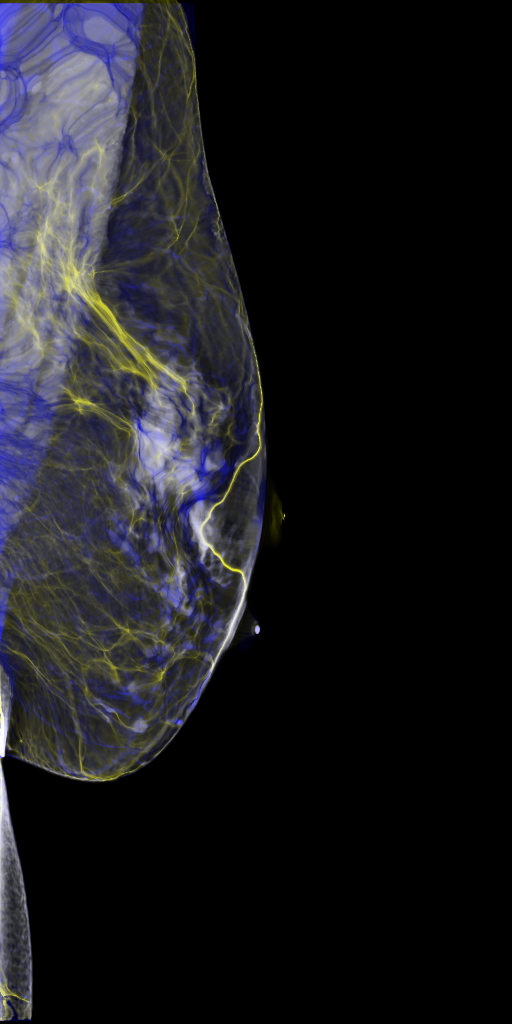}
    \end{minipage}
  }
  \hfill
  \hfill
  \caption{Failure case of MammoRegNet on a challenging mammogram pair with significant anatomical discrepancy. Top row: input moving and fixed images, and registration results using standalone MammoRegNet versus MammoRegNet with pre-applied affine alignment. Bottom row: corresponding overlays (fixed in blue, registered moving in yellow). Despite affine pre-alignment, both variants produce severe unnatural deformations, indicating limitations of the method on outlier cases.}
  \label{fig:mrn2}
\end{figure*}

\end{document}